\documentclass[lettersize,journal]{IEEEtran}
\usepackage{amsmath,amsfonts}
\usepackage{algorithmic}
\usepackage{algorithm}
\usepackage{array}
\usepackage[caption=false,font=normalsize,labelfont=sf,textfont=sf]{subfig}
\usepackage{textcomp}
\usepackage{stfloats}
\usepackage{url}
\usepackage{verbatim}
\usepackage{graphicx}
\usepackage{cite}
\begin{document}

\title{A Survey of Embodied AI: From Simulators to Research Tasks}

\author{Jiafei Duan, Samson Yu, Hui Li Tan, Hongyuan Zhu, Cheston Tan
        % <-this % stops a space
        
\thanks{This work was supported by the Agency for Science, Technology
and Research (A*STAR), Singapore under its AME
Programmatic Funding Scheme (Award \#A18A2b0046) and
the National Research Foundation, Singapore under its NRFISF
Joint Call (Award NRF2015-NRF-ISF001-2541).}
\thanks{J. Duan was with the Nanyang Technological University of Singapore, School of Electrical and Electronics Engineering, Singapore 639798, Singapore (e-mail: duan0038@e.ntu.edu.sg).}
\thanks{S. Yu was with the Singapore University of Technology and Design, Singapore 487372, Singapore (e-mail: samson\_yu@sutd.edu.sg).}
\thanks{H.L. Tan, H. Zhu, and C. Tan are with the Institute for Infocomm Research, A*STAR, Singapore 138632, Singapore (e-mail: \{duan\_jiafei, hltan, zhuh, cheston-tan\}@i2r.a-star.edu.sg).}
% <-this % stops a space
\thanks{\emph{Manuscript accepted December 4, 2021, IEEE-TETCI. \copyright 2021 IEEE. Personal use of this material is permitted. Permission from IEEE must be obtained for all other uses, in any current or future media, including reprinting/republishing this material for advertising or promotional purposes, creating new collective works, for resale or redistribution to servers or lists, or reuse of any copyrighted component of this work in other works.}}}

\markboth{IEEE TRANSACTIONS ON XXXX, VOL. X, NO. X, MM YYYY}%
{Shell \MakeLowercase{\textit{et al.}}: A Sample Article Using IEEEtran.cls for IEEE Journals}


\maketitle

\begin{abstract}
There has been an emerging paradigm shift from the era of ``internet AI'' to ``embodied AI'', where AI algorithms and agents no longer learn from datasets of images, videos or text curated primarily from the internet. Instead, they learn through interactions with their environments from an egocentric perception similar to humans. Consequently, there has been substantial growth in the demand for embodied AI simulators to support various embodied AI research tasks. This growing interest in embodied AI is beneficial to the greater pursuit of Artificial General Intelligence (AGI), but there has not been a contemporary and comprehensive survey of this field. This paper aims to provide an encyclopedic survey for the field of embodied AI, from its simulators to its research. By evaluating nine current embodied AI simulators with our proposed seven features, this paper aims to understand the simulators in their provision for use in embodied AI research and their limitations. Lastly, this paper surveys the three main research tasks in embodied AI -- visual exploration, visual navigation and embodied question answering (QA), covering the state-of-the-art approaches, evaluation metrics and datasets. Finally, with the new insights revealed through surveying the field, the paper will provide suggestions for simulator-for-task selections and recommendations for the future directions of the field.
\end{abstract}


% \fi
\IEEEpeerreviewmaketitle

\section{Introduction}

\IEEEPARstart{R}{ecent} advances in deep learning, reinforcement learning, computer graphics and robotics have garnered growing interest in developing general-purpose AI systems. As a result, there has been a shift from ``internet AI'' that focuses on learning from datasets of images, videos and text curated from the internet, towards ``embodied AI'' which enables artificial agents to learn through interactions with their surrounding environments. Embodied AI is the belief that true intelligence can emerge from the interactions of an agent with its environment \cite{smith2005development}. But for now, embodied AI is about incorporating traditional intelligence concepts from vision, language, and reasoning into an artificial embodiment to help solve AI problems in a virtual environment.  

% Modern techniques in machine learning, computer vision, natural language processing and robotics have achieved great successes in their respective fields, and the resulting applications have enhanced many aspects of technology and human life in general \cite{mnih2013playing,brown2020language,alquraishi2019alphafold}. However, there are still considerable limitations in existing techniques. Typically known as ``weak AI" \cite{nilsson2005human}, existing techniques are confined to pre-defined settings, where the nature of the environment does not change significantly \cite{pfeifer2004embodied}. However, the real world is much more complicated and hence further amplifies the difficulties by vast margins. Despite monumental technological advancements such as the AI systems that can beat most of the world champions in their respective games -- such as chess \cite{campbell2002deep}, Go \cite{silver2017mastering}, and Atari games \cite{mnih2013playing} -- existing AI systems still do not possess such level of effectiveness and sophistication\cite{rubio2019review}.
%four-legged animal.

The growing interest in embodied AI has led to significant progress in embodied AI simulators that aim to faithfully replicate the physical world. These simulated worlds serve as virtual testbeds to train and test embodied AI frameworks before deploying them into the real world. These embodied AI simulators also facilitate the collection of task-based dataset \cite{duan2020actionet,shridhar2020alfred} which are tedious to collect in real-world as it requires an extensive amount of manual labor to replicate the same setting as in the virtual world. 
% \textcolor{blue}{this point about task-based dataset is not well discussed in this survey. if you want to include this point, need to give example of task-based dataset ie include citation + elaborate on why real world collection is difficult.}
%which can be used to tackle task-based learning and hence further bridging the gap between simulation and the real world. 
While there have been several survey papers in the field of embodied AI \cite{pfeifer2004embodied,haugeland1985artificial,Pfeifer2006HowTB}, they are mostly outdated as they were published before the modern deep learning era, which started around 2009 \cite{deng2009imagenet,lecun2015deep,he2016deep,silver2017mastering}.
%as a result of works such as Imagenet \cite{deng2009imagenet}, AlexNet \cite{krizhevsky2012imagenet}, deep learning \cite{lecun2015deep}, ResNet \cite{he2016deep} and AphlaGO \cite{silver2017mastering}.
To the best of our knowledge, there is only one survey paper on the evaluating embodied navigation \cite{anderson2018evaluation} .

\label{sec:connections}
% There is a tight connection between embodied AI simulators and research tasks, as the simulators serve to create ideal virtual testbeds for training and testing of embodied AI frameworks before they are deployed into the physical world. 
%Such simulators are required to accurately replicate the physical world.
%All Embodied AI simulators are not just designed for one specific research task but also an array of general-purpose tasks. This can be reflected from Fig. \ref{fig:connection}, whereby for large- scale simulators such as AI2-THOR, Habitat-Sim, and iGibson \cite{kolve2017ai2, savva2019habitat,xia2018gibson} or smaller scale simulators like VRKitchen and VirtualHome \cite{gao2019vrkitchen,puig2018virtualhome} are all designed for to train agents to perform an array of tasks.
To address the scarcity of contemporary comprehensive survey papers on this emerging field of embodied AI, we propose this survey paper on the field of embodied AI, from its simulators to research tasks. This paper covers the following nine embodied AI simulators that were developed over the past four years:
%This paper will focus on the following nine embodied AI simulators: 
DeepMind Lab \cite{beattie2016deepmind}, AI2-THOR
\cite{kolve2017ai2}, CHALET \cite{yan2018chalet}, VirtualHome \cite{puig2018virtualhome}, VRKitchen \cite{gao2019vrkitchen}, Habitat-Sim \cite{savva2019habitat}, iGibson \cite{xia2020interactive}, SAPIEN \cite{xiang2020sapien}, and ThreeDWorld \cite{gan2020threedworld}. 
The chosen simulators are designed for general-purpose intelligence tasks, unlike game simulators \cite{bellemare2013arcade} which are only used for training reinforcement learning agents. These embodied AI simulators provide realistic representations of the real world in computer simulations, mainly taking the configurations of rooms or apartments that provide some forms of constraint to the environment. The majority of these simulators minimally comprise a physics engine, Python API, and artificial agent that can be controlled or manipulated within the environment.

Embodied AI simulators have given rise to a series of potential embodied AI research tasks, such as \emph{visual exploration}, \emph{visual navigation} and \emph{embodied QA}. We will focus on these three tasks since most existing papers \cite{anderson2018evaluation, ramakrishnan2020exploration, ye2020seeing} in embodied AI either focus on these tasks or make use of modules introduced for these tasks to build models for more complex tasks like audio-visual navigation. These three tasks are also connected in increasing complexity. Visual exploration is a very useful component in visual navigation \cite{chen2019learning, ramakrishnan2020exploration} and used for realistic situations \cite{savinov2018semi, beeching2020learning}, while embodied QA further involves complex QA capabilities that builds on top of vision-and-language navigation. Since language is a common modality and visual QA is a popular task in AI, embodied QA is a natural direction for embodied AI. These three tasks discussed in this paper have been implemented in at least one of the nine proposed embodied AI simulators. However, Sim2Real \cite{kadian2020sim2real, peng2018sim, tobin2017domain} and robotics in the physical world will not be covered in this paper.

These simulators are selected based on the embodied AI simulators from the Embodied AI Challenge in the annual Embodied AI workshop \cite{Blender} at \emph{Conference on Computer Vision and Pattern Recognition }(CVPR). The research tasks are then sourced from direct citations of these simulators.

% To this end, this paper makes three major contributions to the field of embodied AI. Firstly, the paper surveys the \textcolor{blue}{nine} embodied AI simulators and provide insights into the specification and selection process of simulators for research tasks.
% Secondly, the paper provides a systematic look into embodied AI research directions, and the different stages of embodied AI research that are currently available.
% Lastly, the paper establishes the linkages between embodied AI simulators' development and the progress of embodied AI research.

To this end, we will provide a contemporary and comprehensive survey of embodied AI simulators and research through reviewing the development of the field from its simulator to research.
%The contributions of the paper are as follows.
In section \ref{sec:connections}, this paper outlines the overview structure of this survey paper. In section \ref{sec:simulator}, this paper benchmarks nine embodied AI simulators to understand their provision for realism, scalability, interactivity and hence use in embodied AI research.  Finally, based upon the simulators, in section \ref{sec:research}, this paper surveys three main research tasks in embodied AI - visual exploration, visual navigation and embodied question answering (QA), covering the state-of-the-art approaches, evaluation, and datasets. Lastly, this paper will establish interconnections between the simulators, datasets and research tasks and existing challenges in embodied AI simulators and research in section \ref{challenges}. This survey paper provides a comprehensive look into the emerging field of embodied AI and further unveils new insights and challenges of the field. Furthermore, through this paper, we seek to avail AI researchers in selecting the ideal embodied AI simulators for their research tasks of interest.

% In this paper, we will first present the simulators and then present the research that were built upon the simulators.
% The connections between the simulators and research are shown in Fig. \ref{fig:connection}. The simulators are shown in blue and elaborated In section \ref{sec:simulator}. The research tasks are shown in green and elaborated In section \ref{sec:research}. As results of these research tasks, specific datasets are created for the evaluation of the tasks, and are shown in orange.

% \begin{figure*}[thb]
% \begin{minipage}[b]{1.0\linewidth}
%   \centering
%   \centerline{\includegraphics[width=18cm]{connect.jpg}}
%   \caption{Connections between Embodied AI simulators to research. (Top) Nine up-to-date embodied AI simulators. (Middle) The various embodied AI research tasks as a result of the nine embodied AI simulators. The yellow colored research tasks are grouped under the visual navigation category while the rest of the green colored tasks are the other research categories. (Bottom) The evaluation dataset used in the evaluation of the research tasks in one of the nine embodied AI simulators.}
  
% %   \textcolor{blue}{still have the following and previous issues. (1) use same form. use "dataset" or "Dataset" consistently - EQA "dataset" (2) If possible, include the year and reference of dataset. (3) vln is visual-and-language navigation below} \textcolor{red}{Changed}
  
% \label{fig:connection}
% \end{minipage}
% %
% \end{figure*}

\section{Simulators for Embodied AI} \label{sec:simulator}.

% Over the past few years, there has been an emerging trend for Embodied AI simulators, that aims to provide a realistic representation of the real world in a computer simulation. It mainly takes the setting of a room or an apartment, which still provides some form of constraint to the boundary of the environment. Majority of these simulators meet the basic requirement for an Embodied AI simulator such as a physics engine, Python API, and an artificial agent that can be controlled or manipulated in the environment. 

%This paper classifies all the embodied AI simulators that will be described in section \ref{sec:simulator} into two generic classes. They are mainly classified as a game-based embodied AI simulator or a 3D world-based  embodied AI simulator. In the game-based embodied AI simulators, rooms or apartment are all manually designed, sourced, or constructed out of Unity \cite{patil2015cross}. Unity is a cross-platform game engine developed for game developers to create 3D, 2D, virtual reality and even augmented reality games. On the other hand, for the 3D world-based embodied AI simulators, the environment is constructed out of 3D scans of the real-world, hence providing a next level of realism to the simulators. A distinct contrast of the two broad categorizations of the simulators can found in Fig. \ref{fig:environment}.  

% In this section, the backgrounds of the embodied AI simulators will be presented in the supplementary material, and their features will be compared and discussed in section \ref{subsec:briefs}.

%Original
In this section, the backgrounds of the embodied AI simulators will be presented in the supplementary material, and the features of the embodied AI simulators will be compared and discussed in section \ref{subsec:briefs}.

\subsection{Embodied AI Simulators} \label{subsec:briefs}
This section presents the backgrounds of the nine embodied AI simulators: DeepMind Lab, AI2-THOR, SAPIEN, VirtualHome, VRKitchen, ThreeDWorld, CHALET, iGibson, and Habitat-Sim. Readers can refer to the supplementary material for more details on the respective simulators. In this section, the paper will comprehensively compare the nine embodied AI simulators based on seven technical features. Referencing \cite{kolve2017ai2,AllenAct,gan2020threedworld}, these seven technical features are selected as the primary features to evaluate the embodied AI simulator as they cover the essential aspects required to replicate the environment accurately, interactions and state of the physical world, hence providing suitable testbeds for testing intelligence with embodiment.
Referring to Table \ref{tab:features}, the seven features are: Environment, Physics, Object Type, Object Property, Controller, Action, and Multi-Agent.

%(1) Environment: game-based scenes (G) and world-based scenes (W); 
%(2) Physics: basic physics features (B) and advanced physics features (A);
%(3) Object Type: dataset driven environments (D) and object assets driven environments (O);
%(4) Object Property: interact-able objects (I) and multi-state objects (M);
%(5) Controller: direct Python API controller (P), virtual robot controller(R) and virtual reality controller (V);
%(6) Robotic manipulation: navigation (N), atomic action (A) and human-computer interaction (H); and 
%(7) Multi-agent: avatar-based (AT) and user-based (U).

\begin{table*}[!t]

%As shown in Table 1, the benchmark for Embodied AI Simulators for their respective features: 
%\begin{itemize}
%\item Environment: game-based scene (G) and world-based scene (W).
%\item Physics: basic physics features (B) and Advanced physics features (A).
%\item Datasets: dataset driven environment(D) and object assets driven environment(O).
%\item Controller: direct Python API controller (P), virtual robot controller(R) and virtual reality controller (V).
%\item Robotic Manipulation: Navigation (N), atomic action (A) and human-computer interaction (H).
%\item Object Properties: interact-ableobjects (I) and multi-object states (S).
%\item Multi-Agent: avatar-based (AT) and user-based (U).
%\end{itemize}
\caption{Summary of embodied AI simulators. Environment: game-based scene construction (G) and world-based scene construction (W). Physics: basic physics features (B) and advanced physics features (A). Object Type: dataset driven environments (D) and object assets driven environments (O).
Object Property: interact-able objects (I) and multi-state objects (M).
Controller: direct Python API controller (P), virtual robot controller(R) and virtual reality controller (V).
Action: navigation (N), atomic action (A) and human-computer interaction (H). Multi-agent: avatar-based (AT) and user-based (U). The seven features can be further grouped under three secondary evaluation features; realism, scalability and interactivity. }
%\end{itemize} }
\centering
\label{tab1}
\resizebox{\linewidth}{!}{
\begin{tabular}{|p{0.8cm}|p{2cm}|p{1.8cm}|p{1.3cm}|p{1.6cm}|p{1.8cm}|p{1.8cm}| p{1.8cm}|p{1.8cm}|}
\hline
Year & Embodied AI Simulator & Environment \newline (Realism) & Physics \newline (Realism) & Object Type \newline (Scalability)&  Object \newline Property \newline (Interactivity)
&Controller \newline (Interactivity) & Action \newline (Interactivity) & Multi-agent \newline (Interactivity) \\
\hline \hline
2016& DeepMind Lab&
G& 
-&
-&
-&
P, R& 
N& 
-
\\
\hline
2017& AI2-THOR& 
G& 
B& 
O &
I, M&
P, R& 
A, N&
U
\\
\hline
2018& CHALET&
G& 
B& 
O& 
I, M&
P& 
A, N& 
-
\\

\hline
2018& VirtualHome&
G& 
-&
O& 
I, M&
R& 
A, N & 
-
\\
\hline
2019& VRKitchen&
G& 
B& 
O& 
I, M&
P, V& 
A, N, H& 
-
\\
\hline
2019& Habitat-Sim&
W& 
-&
D& 
-&
- &
N& 
-
\\
\hline
2019& iGibson&
W& 
B&
D& 
I&
P, R& 
A, N& 
U 

\\

\hline
2020& SAPIEN&
G& 
B& 
D& 
I, M&
P, R& 
A, N & 
-
\\
\hline

2020& ThreeDWorld&
G& 
B, A&
O& 
I&
P, R, V& 
A, N, H& 
AT\\
\hline

% \multicolumn{6}{@{}l}{$^*$TAR--True Acceptance Rate\qquad 
% $^{\#}$ FAR--False Acceptance Rate}
\end{tabular}}
\label{tab:features}
\end{table*}

\begin{table*}[!t]
\caption{Comparison of embodied ai simulators in terms of environment configuration, simulation engine, technical specification, and rendering performance.}
%\end{itemize} }
\centering
\resizebox{\linewidth}{!}{
\begin{tabular}{|l|l|l|l|l|}
\hline
Embodied AI Simulator & Environment Configuration& Simulation Engine & Technical Specification&  Rendering Performance \\
\hline \hline
DeepMind Lab&
Customized environment&
Quake II Arena Engine& 
6-core Intel Xeon CPU and an NVIDIA Quadro K600 GPU&
158 fps$/$thread\\
\hline
AI2-THOR& 
120 rooms, 4 categories& 
Unity 3D Engine& 
Intel(R) Xeon(R) CPU E5-2620 v4 and NVIDIA Titan X &
240 fps$/$thread
\\
\hline
CHALET&
58 rooms, 10 houses& 
Unity 3D Engine& 
-& 
-
\\

\hline
VirtualHome&
6 apartments with multiple jointed rooms& 
Unity 3D Engine&
-& 
Customized frame rate
\\
\hline
VRKitchen&
16 kitchens& 
Unreal Engine 4& 
Intel(R) Core(TM) i7-7700K processor and NVIDIA Titan X & 
15 fps$/$thread

\\
\hline
Habitat-Sim&
Mutiple datasets& 
-&
Xeon E5-2690 v4 CPU and Nvidia Titan Xp GPU& 
10,000 fps$/$thread
\\
\hline
iGibson&
Gibson V1& 
-&
Modern GPU
& 
1000 fps$/$thread

\\

\hline
SAPIEN&
Customized environment& 
PhysX Physical engine and ROS& 
Intel i7-8750 CPU and an Nvidia GeForce RTX 2070 GPU& 
700 fps$/$thread
\\
\hline

ThreeDWorld&
Customized environment& 
Unity 3D Engine&
Intel i7-7700K GPU: NVIDIA GeForce GTX 1080& 
168 fps$/$thread\\
\hline

% \multicolumn{6}{@{}l}{$^*$TAR--True Acceptance Rate\qquad 
% $^{\#}$ FAR--False Acceptance Rate}
\end{tabular}}
\label{tab:quantitative}
\end{table*}

\begin{figure}[htb]
\begin{minipage}[b]{1.0\linewidth}
  \centering
  \includegraphics[width=\linewidth]{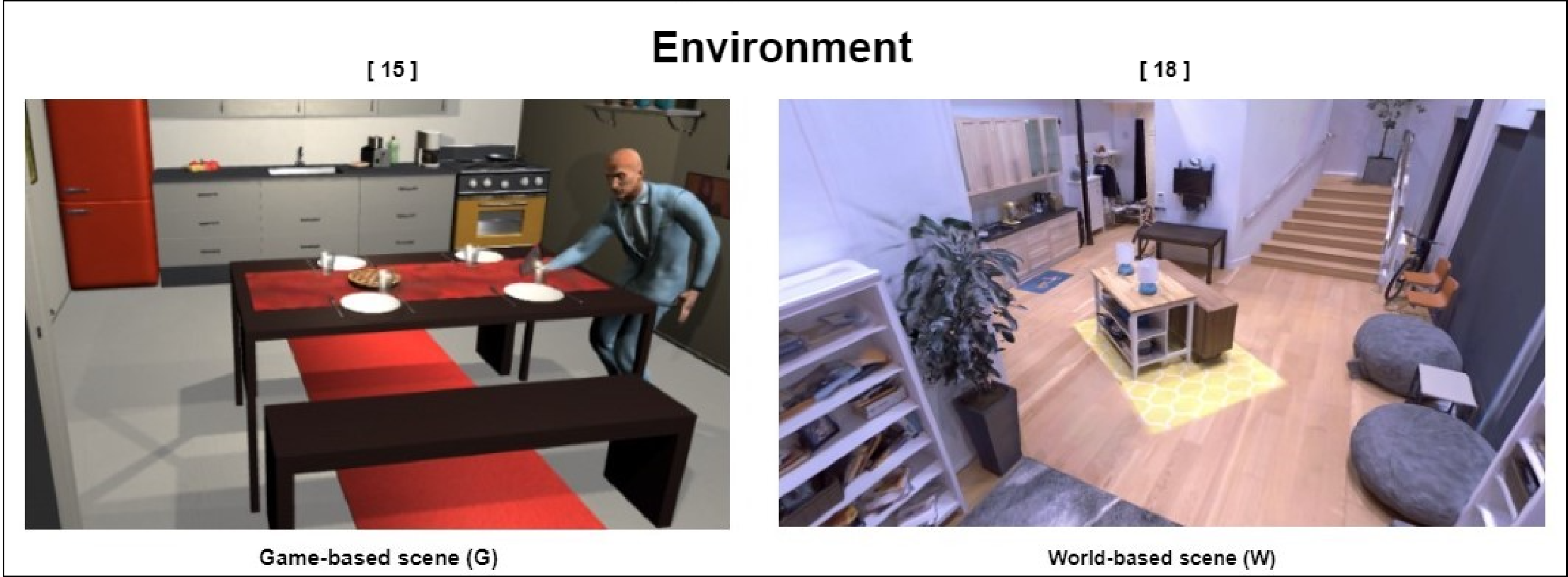}
  \caption{Comparison between game-based scene (G) and world-based scene (W). The game-based scene (G) focuses on environment that are constructed from 3D object assets, while the world-based scene (W) are constructed based off real-world scans of the environment. }
\label{fig:environment}
\end{minipage}
\end{figure}

 %The realism of a 3D environment is dependent on the method of constructing the environment. 
%We study the realism of the 3D environments based on two common methods of constructing the environments, i.e.,
\textbf{Environment}: There are two main methods of constructing the embodied AI simulator environment: game-based scene construction (G) and world-based scene construction (W). 
Referring to Fig. \ref{fig:environment}, the game-based scenes are constructed from 3D assets, while world-based scenes are constructed from real-world scans of the objects and the environment.
A 3D environment constructed entirely out of 3D assets often has built-in physics features and object classes that are well-segmented when compared to a 3D mesh of an environment made from real-world scanning. 
The clear object segmentation for the 3D assets makes it easy to model them as articulated objects with movable joints, such as the 3D models provided in PartNet \cite{mo2019partnet}. In contrast, the real-world scans of environments and objects provide higher fidelity and more accurate representation of the real-world, facilitating better transfer of agent performance from simulation to the real world.
As observed in Table \ref{tab1}, most simulators other than Habitat-Sim and iGibson have game-based scenes, since significantly more resources are required for world-based scene construction.  

\begin{figure}[t]
\begin{center}
\includegraphics[width=\linewidth]{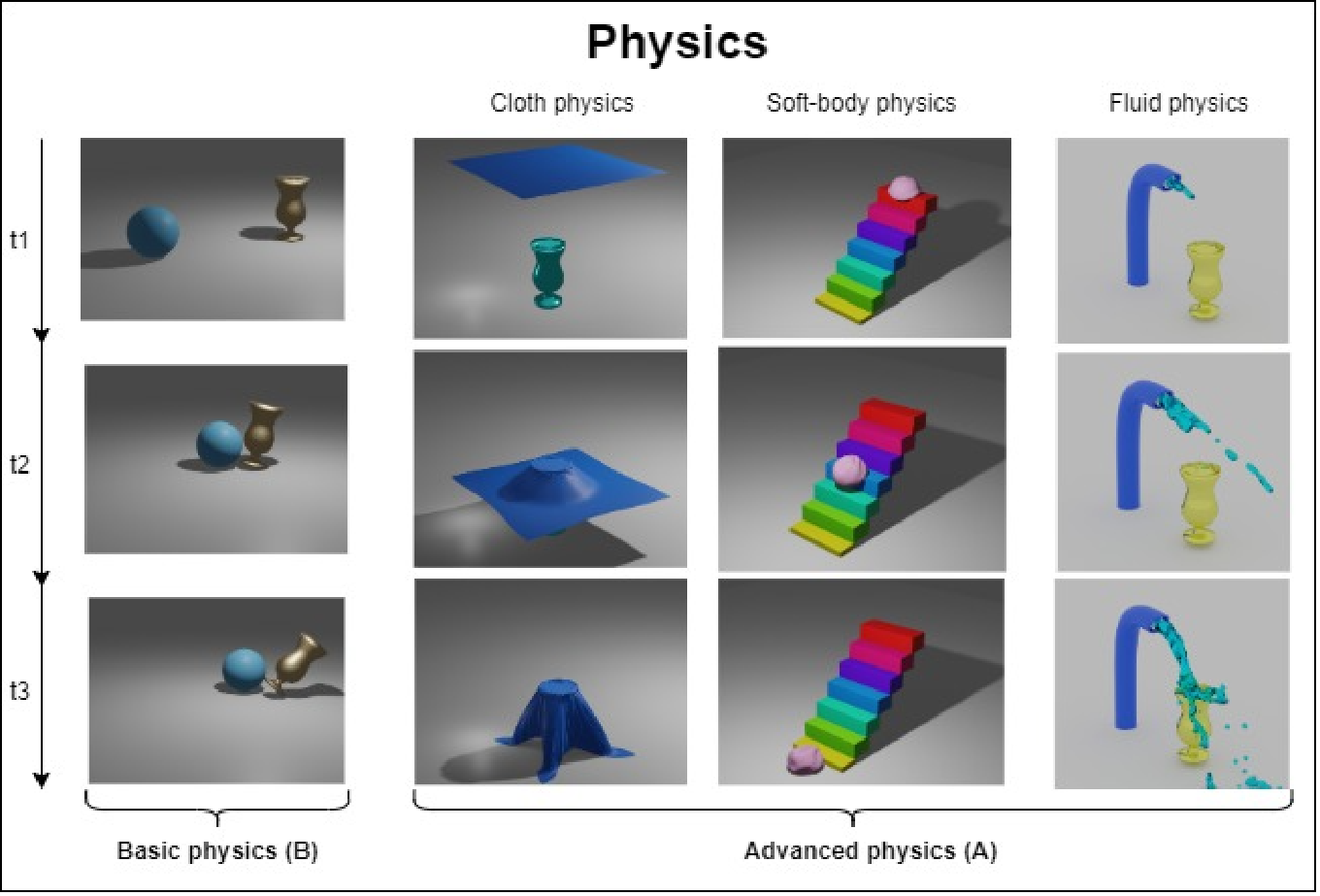}
\end{center}
  \caption{Comparison between basics physics features such as rigid-body and collision (B) and advanced physics features (A) which includes cloth, soft-body, and fluid physics.}
\label{fig:Physics}
\end{figure}

\textbf{Physics}: A simulator has to construct not only realistic environments but also realistic interactions between agents and objects or objects and objects that model real-world physics properties. We study the simulators' physics features, which we broadly classify into basic physics features (B) and advanced physics features (A).
Referring to Fig. \ref{fig:Physics}, basic physics features include collision, rigid-body dynamics, and gravity modelling while advanced physics features include cloth, fluid, and soft-body physics. As most embodied AI simulators construct game-based scenes with in-built physics engines, they are equipped with the basic physics features. On the other hand, for simulators like ThreeDWorld, where the goal is to understand how the complex physics environment can shape the decisions of the artificial agent in the environment, they are equipped with more advanced physics capabilities. For simulators that focus on interactive navigation-based tasks, basic physics features are generally sufficient. 

\begin{figure*}[thb]
\begin{minipage}[b]{1.0\linewidth}
  \centering
  \includegraphics[width=\linewidth]{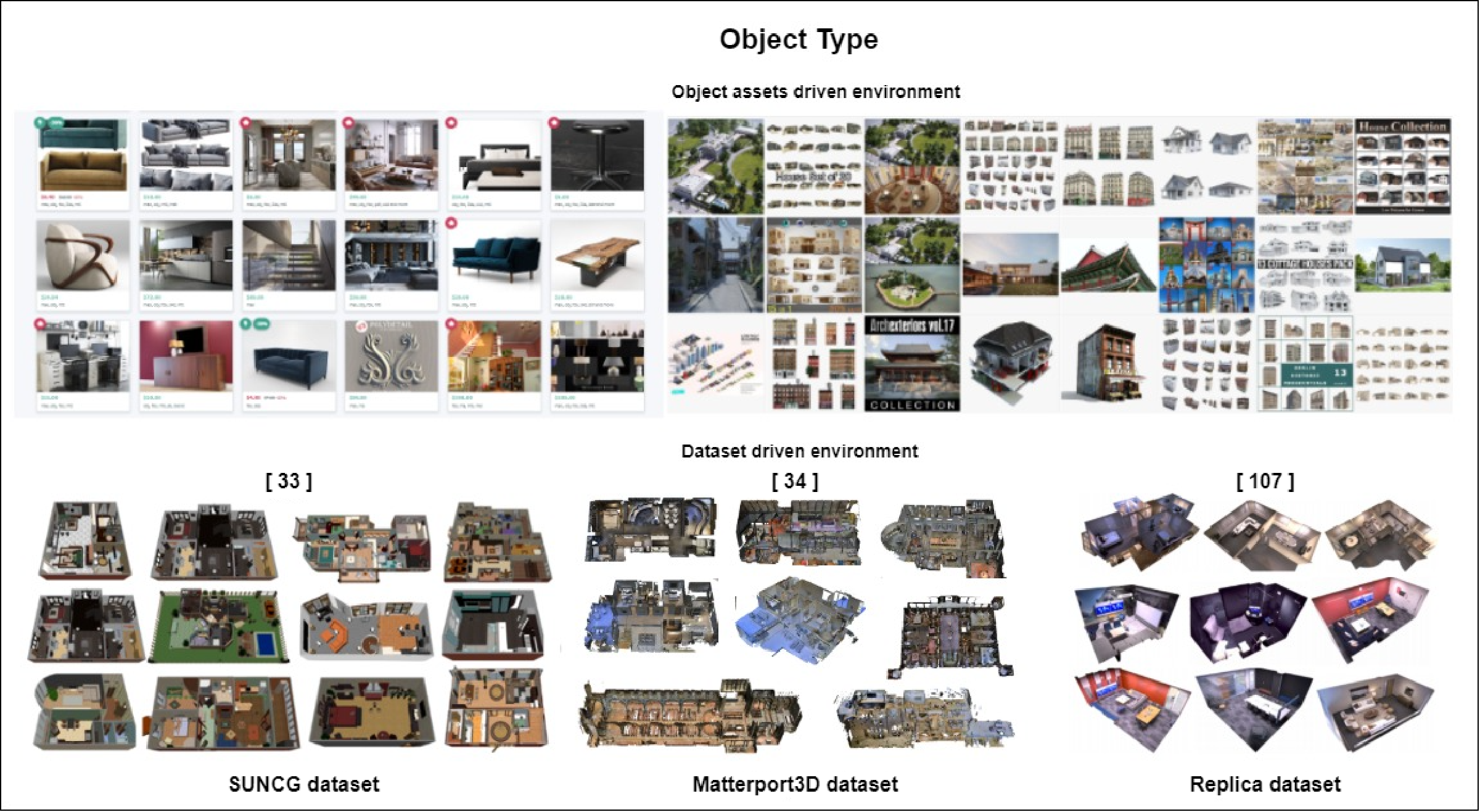}
  \caption{Comparison between dataset driven environment (D) which are constructed from 3D objects datasets and object assets driven environment (O) are constructed based 3D objects obtain from the assets market.}
\label{fig:ObjectType}
\end{minipage}
\end{figure*}

\textbf{Object Type}:
As shown in Fig. \ref{fig:ObjectType}, there are two main sources for objects that are used to create the simulators.
The first type is the dataset driven environment, where the objects are mainly from existing object datasets such as the SUNCG \cite{song2016ssc} dataset, the Matterport3D dataset \cite{chang2017matterport3d} and the Gibson dataset \cite{xia2018gibson}. The second type is the asset driven environment, where the objects are from the net such as the Unity 3D game asset store. A difference between the two sources is the sustainability of the object dataset. The dataset driven objects are more costly to collect than the asset driven objects, as anyone can contribute to the 3D object models online. However, it is harder to ensure the quality of the 3D object models in the asset driven objects than in the dataset driven objects. Based on our review, the game-based embodied AI simulators are more likely to obtain their object datasets from asset stores, whereas the world-based simulators tend to import their object datasets from existing 3D object datasets.

\textbf{Object Property}:
%We look into the interactivity of an object in the virtual environment.
Some simulators only enable objects with basic interactivity such as collision.
Advanced simulators enable objects with more fine-grained interactivity such as multiple-state changes. 
For instance, when an apple is sliced, it will undergo a state change into apple slices. Hence, we categorize these different levels of object interaction into simulators with interact-able objects (I) and multiple-state objects (M).
Referring to Table \ref{tab1}, a few simulators, such as AI2-THOR and VRKitchen, enable multiple state changes, providing a platform for understanding how objects will react and change their states when acted upon in the real world.

\begin{figure*}[thb]
\begin{minipage}[b]{1.0\linewidth}
  \centering
  \includegraphics[width=\linewidth]{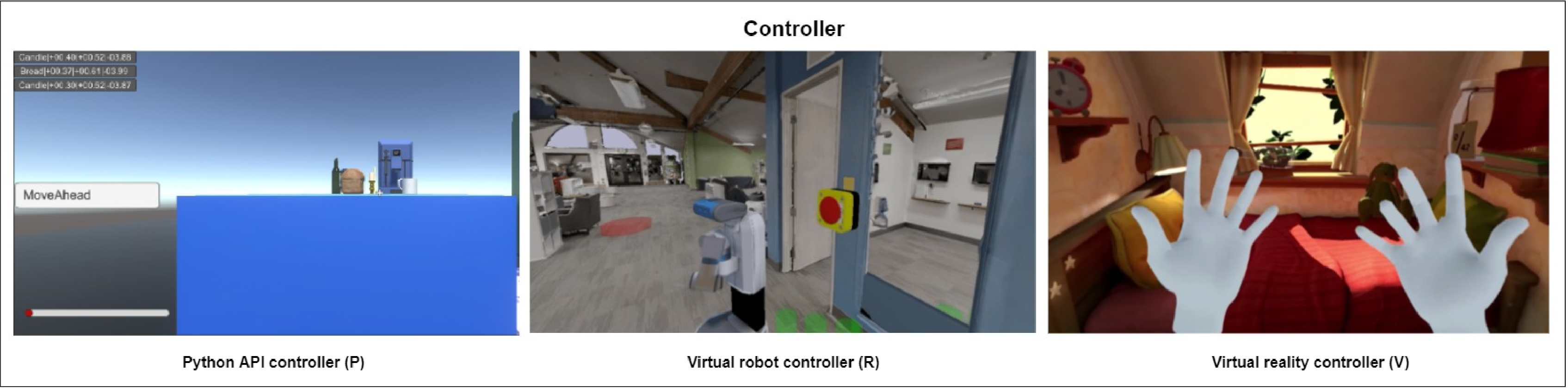}
  \caption{Comparison between direct Python API controller  (P), robotics embodiment (R) which refers to real-world robots with a virtual replica and lastly the virtual reality controller (V).}
\label{fig:Interaction}
\end{minipage}
\end{figure*}

\textbf{Controller}: Referring to Fig. \ref{fig:Interaction}, there are different types of controller interface between the user and simulator, from direct Python API controller (P) and virtual robot controller(R) to virtual reality controller (V). Robotics embodiment allows for virtual interaction of existing real-world robots such as Universal Robot 5 (UR5) and TurtleBot V2, and can be controlled directly using a ROS interface. The virtual reality controller interfaces provide more immersive human-computer interaction and facilitate deployment using their real-world counterparts.
For instance, simulators such as iGibson and AI2-THOR, which are primarily designed for visual navigation, are also equipped with virtual robot controllerfor ease of deployment in their real-world counterparts such as iGibson’s Castro \cite{habitat2020sim2real} and RoboTHOR \cite{deitke2020RoboTHOR} respectively. 

\textbf{Action}: There are differences in the complexity of an artificial agent's action capabilities in the embodied AI simulator, ranging from being only able to perform primary navigation manoeuvers to higher-level human-computer actions via virtual reality interfaces. This paper classifies them into three tiers of robotics manipulation: navigation (N), atomic action (A) and human-computer interaction (H). Navigation is the lowest tier and is a common feature in all embodied AI simulators \cite{batra2020objectnav}. It is defined by the agent's capability of navigating around its virtual environment. Atomic action provides the artificial agent with a means of performing basic discrete manipulation to an object of interest and is found in most embodied AI simulators. Human-computer interaction is the result of the virtual reality controller as it enables humans to control virtual agents to learn and interact with the simulated world in real time \cite{gao2019vrkitchen}.
%Different complexities of tasks require different Actions. 
Most of the larger-scale navigation-based simulators, such as AI2-THOR, iGibson and Habitat-Sim, tend to have navigation, atomic action and ROS \cite{kolve2017ai2,xia2018gibson,savva2019habitat} which enable them to provide better control and manipulation of objects in the environment while performing tasks such as Point Navigation or Object Navigation. On the other hand, simulators such as ThreeDWorld and VRKitchen \cite{gan2020threedworld,gao2019vrkitchen} fall under the human-computer interaction category as they are constructed to provide a highly realistic physics-based simulation and multiple state changes. This is only possible with human-computer interaction as human-level dexterity is needed when interacting with these virtual objects.

\textbf{Multi-agent}: Referring to Table \ref{tab1}, only a few simulators, such as AI2-THOR, iGibson and ThreeDWorld, are equipped with multi-agent setup, as current research involving multi-agent reinforcement learning is scarce.
%Only three embodied AI simulators support the multi-agent setting. 
In general, the simulators need to be rich in object content before there is any practical value of constructing such multi-agent features used for both adversarial and collaborative training \cite{jaincordial,JainWeihs2019TwoBody} of artificial agents. As a result of this lack of multi-agent supported simulators, there have been fewer research tasks that utilize the multi-agent feature in these embodied AI simulators.

For multi-agent reinforcement learning based training, they are still currently being done in OpenAI Gym environments \cite{brockman2016openai} . There are two distinct multi-agent settings. The first is the avatar-based (AT) multi-agents in ThreeDWorld \cite{gan2020threedworld} that allows for interaction between artificial agents and simulation avatars. The second is the user-based (U) multi-agents in AI2-THOR \cite{kolve2017ai2} which can take on the role of a dual learning network and learn from interacting with other artificial agents in the simulation to achieve a common task \cite{jain2019two}.
% \textcolor{blue}{the description for user based does not seem to reflect "user based" well.}

% \begin{figure}[t]
% \begin{center}
% \includegraphics[width=\linewidth]{dataexample.png}
% \end{center}
%   \caption{ Top row: visual data attributes for one example frame comprises of RGB, object segmentation, optical flow, depth, and surface normal vector. Bottom three rows: example frames from the three physical interactions.}
% \label{fig:long3}
% \end{figure}

% \begin{figure}[t]
% \begin{minipage}[b]{1.0\linewidth}
%   \centering
%   \centerline{\includegraphics[width=18cm]{physics.png}}
%   \caption{Comparison between basics physics features such as rigid-body and collision (B) and advanced physics features (A) which includes cloth, soft-body, and fluid physics.}
% \label{fig:Physics}
% \end{minipage}
% %
% \end{figure}

% \begin{figure*}[thb]
% \begin{minipage}[b]{1.0\linewidth}
%   \centering
%   \centerline{\includegraphics[width=\linewidth]{connection.png}}
%   \caption{Connections between Embodied AI simulators to research. (Top) Nine up-to-date embodied AI simulators. (Middle) The various embodied AI research tasks as a result of the nine embodied AI simulators. The yellow colored research tasks are grouped under the visual navigation category while the rest of the green colored tasks are the other research categories. (Bottom) The evaluation dataset used in the evaluation of the research tasks in one of the nine embodied AI simulators.}
  
% %   \textcolor{blue}{still have the following and previous issues. (1) use same form. use "dataset" or "Dataset" consistently - EQA "dataset" (2) If possible, include the year and reference of dataset. (3) vln is visual-and-language navigation below} \textcolor{red}{Changed}
  
% \label{fig:connection}
% \end{minipage}
% %
% \end{figure*}

\subsection{Comparison of Embodied AI Simulators}

Constructed on the seven features and a study from the Allen Institute of Artificial Intelligence \cite{AllenAct} on embodied AI, we propose a secondary set of evaluation features for the simulators. It comprises of three key features: \emph{realism}, \emph{scalability} and \emph{interactivity} as shown in Table \ref{tab:features}. 
The realism of the 3D environments can be attributed to the \emph{environment} and \emph{physics} of the simulators. 
The environment models the real world's physical appearance while the physics models the complex physical properties within the real world.
Scalability of the 3D environments can be attributed to the \emph{object type}. The expansion can be done via collecting more 3D scans of the real world for the dataset driven objects or purchasing more 3D assets for the asset driven objects. \emph{Interactivity} is attributed to \emph{object property}, \emph{controller}, \emph{action} and \emph{multi-agent}.

Based on the secondary evaluation features of embodied AI simulators, the seven primary features from the Table \ref {tab:features} and the Fig. \ref{fig:connection}, simulators which possess all of the above three secondary features (e.g. AI2-THOR, iGibson and Habitat-Sim) are more well-received and widely used for a diverse range of embodied AI research tasks. Furthermore, a comprehensive quantitative comparison is made for all the embodied AI simulators to compare the environment configuration and the technical performance of each simulator. The \textbf{environment configuration} feature is very much dependent on the applications suggested by the creators of the simulators, while other features like \textbf{technical specification} and \textbf{rendering performance} are largely due to the \textbf{simulation engine} used for its creation. AI2-THOR has the largest environment configurations compared to the other simulators, while Habitat-Sim and iGibson are the top two performers in graphic rendering performance. This benchmark of quantitative performance shown in Table \ref{tab:quantitative} further demonstrates the superiority and complexity of these three embodied AI simulators. These comparisons of the embodied AI simulators further have reinforced the importance of the seven primary evaluation metrics and the three secondary evaluations that the paper has established to help select the ideal simulator for the research task.

 \section{Research in Embodied AI} \label{sec:research}

\begin{table*}[!t]

\caption{\label{tasks} Summary of embodied AI research tasks.
Evaluation Metric: Amount of targets visited (ATV),  downstream tasks (D), success weighted by path length (SPL), success rate (SR), path length ratio (PLR), oracle success rate (OSR), trajectory/episode length (TL / EL), distance to success / navigation error (DTS / NE / $d_T$), goal progress (GP / $d_\Delta$), oracle path success rate (OPSR), smallest distance to target at any point in an episode ($d_{min}$), percentage of episodes agent ends navigation for answering before max episode length ($\%stop$), percentage of questions agent terminates in the room containing the target object ($\%r_T$), percentage of questions where the agent enters the room containing the target oject at least once ($\%r_e$), Intersection over Union for target object (IoU), hit accuracy based on IoU ($h_T$), mean rank of the ground-truth answer in QA predictions (MR) and QA accuracy (Acc).}
%\end{itemize} }
\centering
\resizebox{\linewidth}{!}{
\begin{tabular}{|c|c|c|c|c|c|c|}
\hline
Task & Method / Category & Publication & Year & Simulator & Dataset & Evaluation Metric \\
\hline \hline
Visual Exploration & Curiosity & Chaplot et al. \cite{chaplot2020semantic} & 2020 & Habitat-Sim & Matterport3D, Gibson V1 & ATV \\
\cline{2-7} & Coverage & Chaplot et al. \cite{chaplot2020learning} & 2020 & Habitat-Sim & Matterport3D, Gibson V1 & ATV, D \\
\cline{2-7} & Reconstruction & Ramakrishnan et al. \cite{ramakrishnan2020occupancy} & 2020 & Habitat-Sim & Matterport3D, Gibson V1 & ATV, D \\
\cline{3-7} & & Ramakrishnan et al. \cite{ramakrishnan2020exploration} & 2020 & Habitat-Sim & Matterport3D & ATV, D \\
\cline{3-7} & & Narasimhan et al. \cite{narasimhan2020seeing} & 2020 & Habitat-Sim & Matterport3D & ATV, D \\
\hline

Visual Navigation & Point Navigation & Wijmans et al. \cite{wijmans2019dd} & 2019 & Habitat-Sim & Matterport3D, Gibson V1 & SPL, SR\\
\cline{3-7} & & Georgakis et al. \cite{georgakis2019simultaneous} & 2019 & Habitat-Sim & Matterport3D & SR, PLR \\
\cline{3-7} & & Ye et al. \cite{ye2020auxiliary} & 2020 & Habitat-Sim & Gibson V1 & SPL, SR \\
\cline{3-7} & & Chaplot et al. \cite{chaplot2020learning} & 2020 & Habitat-Sim & Matterport3D, Gibson V1 & SPL, SR \\
\cline{3-7} & & Ramakrishnan et al. \cite{ramakrishnan2020occupancy} & 2020 & Habitat-Sim & Matterport3D, Gibson V1 & SPL, SR \\
\cline{3-7} & & Ramakrishnan et al. \cite{ramakrishnan2020exploration} & 2020 & Habitat-Sim & Matterport3D & SPL \\
\cline{3-7} & & Narasimhan et al. \cite{narasimhan2020seeing} & 2020 & Habitat-Sim & Matterport3D & SPL, SR \\
\cline{3-7} & & Claudia, et al. \cite{perez2020robot} & 2020 & iGibson & Gibson V1 & SR \\

\cline{2-7} & Object Navigation & Wortsman et al. \cite{wortsman2019learning} & 2019 & AI2-THOR & - & SPL, SR \\
\cline{3-7} & & Campari et al. \cite{campari2020exploiting} & 2020 & Habitat-Sim & Matterport3D & SPL, SR, DTS \\
\cline{3-7} & & Du et al. \cite{du2020learning} & 2020 & AI2-THOR & - & SPL, SR \\
\cline{3-7} & & Chaplot et al. \cite{chaplot2020object} & 2020 & Habitat-Sim & Matterport3D, Gibson V1 & SPL, SR, DTS \\
\cline{3-7} & & Shen et al. \cite{shenigibson} & 2020 & iGibson & Gibson V1 & SR \\
\cline{3-7} & & Wahid et al. \cite{wahid2020learning} & 2020 & - & Gibson V1 & SPL, SR \\

\cline{2-7} & Navigation with Priors & Yang et al. \cite{yang2018visual} & 2020 & AI2-THOR & - & SPL, SR \\

\cline{2-7} & Vision-and-Language Navigation & Anderson et al. \cite{anderson2018vision} & 2018 & - & Room-to-Room & SR, OSR, TL, NE \\
\cline{3-7} & & Zhu et al. \cite{zhu2020vision} & 2020 & - & Room-to-Room & SPL, SR, OSR, TL, NE \\
\cline{3-7} & & Zhu et al. \cite{zhu2020vision2} & 2020 & - & Cooperative Vision-and-Dialog Navigation & SR, OSR, GP, OPSR \\
\hline

Embodied Question Answering & Question Answering & Das et al. \cite{das2018embodied} & 2018 & - & EQA & $d_T$, $d_\Delta$, $d_{min}$, $\%r_T$, $\%r_e$, $\%stop$, MR \\
\cline{3-7} & & Das et al. \cite{das2018neural} & 2018 &-& EQA & $d_T$, $d_\Delta$, Acc \\
\cline{2-7} & Multi-target Question Answering & Yu et al. \cite{MultiTarget;YuCGBBB19} & 2019 &-& MT-EQA & $d_T$, $d_\Delta$, $\%r_T$, $\%stop$, IoU, $h_T$, Acc \\
\cline{2-7} & Interactive Question Answering & Gordon et al. \cite{gordon2018iqa} & 2018 & AI2-THOR & IQUAD V1 & EL, Acc \\
\cline{3-7} & & Tan et al. \cite{MultiAgentVQA;TanXLGS20} & 2020 & AI2-THOR & IQUAD V1 & EL, Acc \\
\hline

% \multicolumn{6}{@{}l}{$^*$TAR--True Acceptance Rate\qquad 
% $^{\#}$ FAR--False Acceptance Rate}
\end{tabular}}
\label{tab:tasks}
\end{table*}

In this section, we discuss the various embodied AI research tasks that depend on the nine embodied AI simulators surveyed in the previous section.
%Research motivations \textcolor{red}{not sure wht this paragraph is for.}
There are multiple motivations for the recent increase in embodied AI research. From a cognitive science and psychology perspective, the embodiment hypothesis \cite{smith2005development} suggests that intelligence arises from interactions with an environment and as a result of sensorimotor activity \cite{nguyen2021sensorimotor}. Intuitively, humans do not learn solely through the ``internet AI" paradigm where most experiences are randomized and passive (i.e. externally curated). Humans also learn through active perception, movement, interaction and communication. From an AI perspective, current research tasks in embodied AI allows for greater generalization to unseen environments \cite{chaplot2020learning} for robotic functions like mapping and navigation and greater robustness to sensor noise as compared to classical methods due to the learning involved. Embodied AI also enables flexibility and possibly greater performance since various modalities like depth, language \cite{zhu2020vision} and audio \cite{chen2020soundspaces} can be easily integrated through learning-based approaches.

\begin{figure}[htb]
\begin{minipage}[b]{1.0\linewidth}
  \centering
  \centerline{\includegraphics[width=\linewidth]{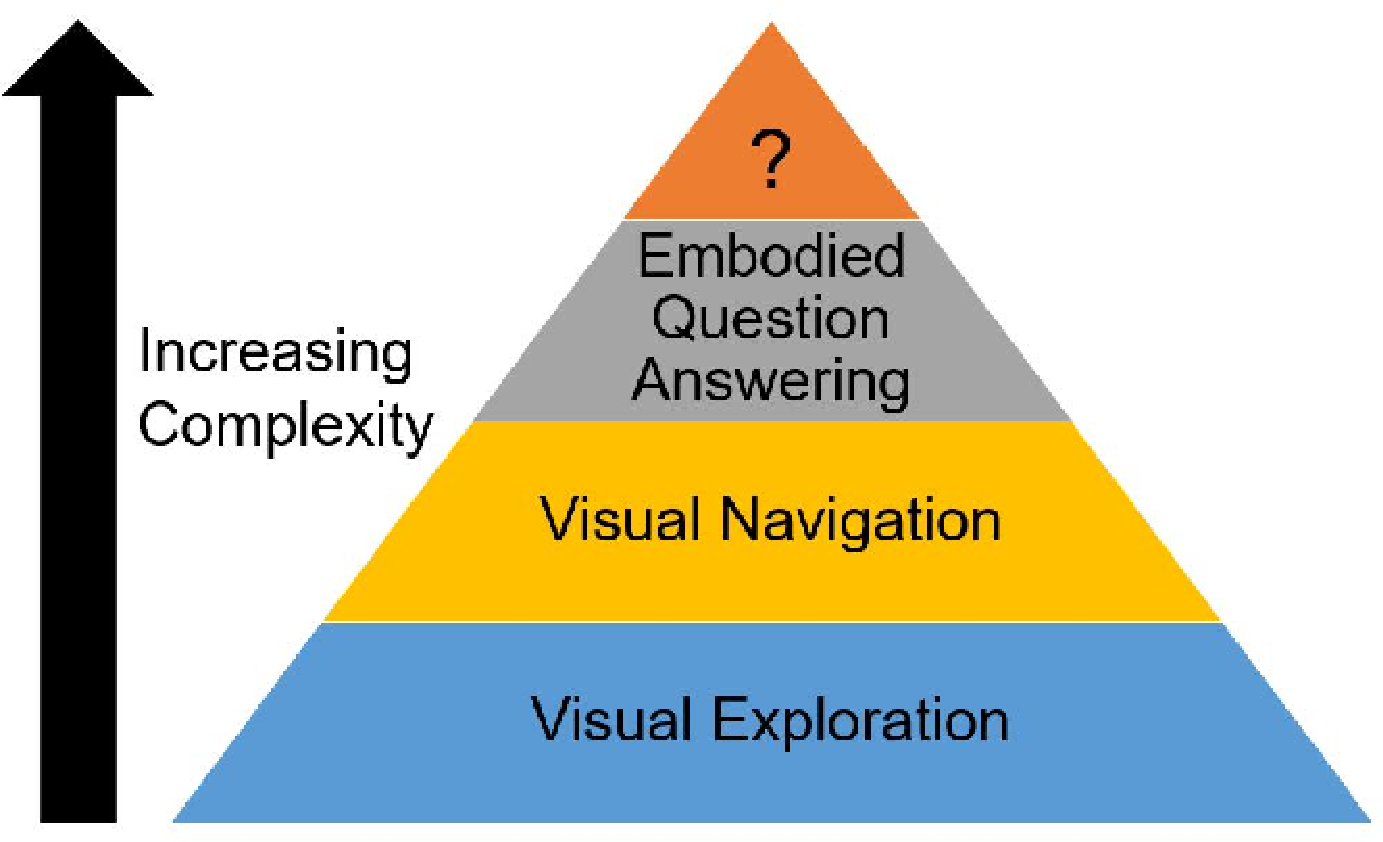}}
  \caption{%Pyramid of embodied AI research, this is a hierarchical map of the different categories of tasks for embodied AI research.
  A pyramid hierarchical structure of the various embodied AI research tasks with increasing complexity of tasks.}
\label{fig:pyramid}
\end{minipage}
\end{figure}

The three main types of embodied AI research tasks are
%There have been various research directions on recent works done with embodied AI simulators. 
\emph{visual exploration}, \emph{visual navigation} and \emph{embodied QA}. We will focus on these three tasks since most existing papers in embodied AI either focus on these tasks or make use of modules introduced for these tasks to build models for more complex tasks like audio-visual navigation. The tasks increase in complexity as it advances from exploration to QA. We will start with the visual exploration before moving to visual navigation and finally embodied QA. Each of these tasks makes up the foundation for the next task(s), forming a pyramid structure of embodied AI research tasks as shown in Fig. \ref{fig:pyramid}, further suggesting a natural direction for embodied AI.
We will highlight important aspects for each task, starting with the summary, the methodologies, evaluation metrics, to the datasets. These task details are found in Table \ref{tasks}.

\subsection{Visual Exploration}
In visual exploration \cite{pathak2017curiosity, chen2019learning}, an agent gathers information about a 3D environment, typically through motion and perception, to update its internal model of the environment \cite{ramakrishnan2020exploration, anderson2018evaluation}, which might be useful for downstream tasks like visual navigation \cite{gupta2017unifying, savinov2018semi, chen2019learning}. The aim is to do this as efficiently as possible (e.g. with as few steps as possible). The internal model can be in forms like a topological graph map \cite{beeching2020learning}, semantic map \cite{narasimhan2020seeing}, occupancy map \cite{ramakrishnan2020occupancy} or spatial memory \cite{gupta2017cognitive, henriques2018mapnet}. These map-based architectures can capture geometry and semantics, allowing for more efficient policy learning and planning \cite{ramakrishnan2020occupancy} as compared to reactive and recurrent neural network policies \cite{fang2019scene}. Visual exploration is usually either done before or concurrently with navigation tasks. In the first case, visual exploration builds the internal memory as priors that are useful for path-planning in downstream navigation tasks. The agent is free to explore the environment within a certain budget (e.g. limited number of steps) before the start of navigation \cite{anderson2018evaluation}. In the latter case, the agent builds the map as it navigates an unseen test environment \cite{mezghani2020learning, georgakis2019simultaneous, mishkin2019benchmarking}, which makes it more tightly integrated with the downstream task. In this section, we build upon existing visual exploration survey papers \cite{ramakrishnan2020exploration, chen2019learning} to include more recent works and directions.

In classical robotics, exploration is done through passive or active simultaneous localisation and mapping (SLAM) \cite{chen2019learning, ramakrishnan2020occupancy} to build a map of the environment. This map is then used with localization and path-planning for navigation tasks. SLAM is very well-studied \cite{cadena2016past}, but the purely geometric approach has room for improvements. Since they rely on sensors, they are susceptible to measurement noise \cite{chen2019learning} and would need extensive fine-tuning. On the other hand, learning-based approaches that typically use RGB and/or depth sensors are more robust to noise \cite{chaplot2020learning, chen2019learning}. Furthermore, learning-based approaches in visual exploration allow an artificial agent to incorporate semantic understanding (e.g. object types in the environment) \cite{ramakrishnan2020occupancy} and generalize its knowledge of previously seen environments to help with understanding novel environments in an unsupervised manner. This reduces reliance on humans and thus improves efficiency. 

Learning to create useful internal models of the environment in the form of maps can improve the agent’s performance \cite{ramakrishnan2020occupancy}, whether it is done before (i.e. unspecified downstream tasks) or concurrently with downstream tasks. Intelligent exploration would also be especially useful in cases where the agent has to explore novel environments that dynamically unfold over time \cite{ramakrishnan2019emergence}, such as rescue robots and deep-sea exploration robots.

\subsubsection{Approaches}
In this section, the non-baseline approaches in visual exploration are typically formalized as partially observed Markov decision processes (POMDPs) \cite{lovejoy1991survey}. A POMDP can be represented by a 7-tuple $(S,A,T,R,\Omega,O,\gamma)$ with state space $S$, action space $A$, transition distribution $T$, reward function $R$, observation space $\Omega$, observation distribution $O$ and discount factor $\gamma \in [0,1]$. In general, these approaches are viewed as a particular reward function in the POMDP \cite{ramakrishnan2020exploration}.

\textbf{Baselines}. Visual exploration has a few common baselines \cite{ramakrishnan2020exploration}. For \emph{random-actions} \cite{savva2019habitat}, the agent samples from a uniform distribution over all actions. For \emph{forward-action}, it always chooses the forward action. For \emph{forward-action+}, the agent chooses the forward action, but turns left if it collides. For \emph{frontier-exploration}, it visits the edges between free and unexplored spaces iteratively using a map \cite{yamauchi1997frontier, chen2019learning}.

\textbf{Curiosity}. In the \emph{curiosity} approach, the agent seeks states that are difficult to predict. The prediction error is used as the reward signal for reinforcement learning \cite{burda2018large, houthooft2016vime}. This focuses on intrinsic rewards and motivation rather than external rewards from the environment, which is beneficial in cases where external rewards are sparse \cite{pathak2019self}. There is usually a forward-dynamics model that minimises the loss: $L(\hat{s}_{t+1},s_{t+1})$. In this case, $\hat{s}_{t+1}$ is the \emph{predicted} next state if the agent takes action $a_t$ when it is in state $s_t$, while $s_{t+1}$ is the \emph{actual} next state that the agent will end up in. Practical considerations for curiosity have been listed in recent work \cite{burda2018large}, such as using Proximal Policy Optimization (PPO) for policy optimisation. Curiosity has been used to generate more advanced maps like semantic maps in recent work \cite{chaplot2020semantic}. Stochasticity poses a serious challenge in the curiosity approach, since the forward-dynamics model can exploit stochasticity \cite{burda2018large} for high prediction errors (i.e. high rewards). This can arise due to factors like the ``noisy-TV" problem or noise in the execution of the agent's actions \cite{pathak2019self}. One proposed solution is the use of an inverse-dynamics model \cite{pathak2017curiosity} that estimates the action $a_{t-1}$ taken by the agent to move from its previous state $s_{t-1}$ to its current state $s_t$, which helps the agent understand what its actions can control in the environment. While this method attempts to address stochasticity due to the environment, it may be insufficient in addressing stochasticity that results from the agent's actions. One example is the agent's use of a remote controller to randomly change TV channels, allowing it to accumulate rewards without progress. To address this more challenging issue specifically, there have been a few methods proposed recently. Random Distillation Network \cite{burda2018exploration} is one method that predicts the output of a randomly initialized neural network, as the answer is a deterministic function of its inputs. Another method is Exploration by Disagreement \cite{pathak2019self}, where the agent is incentivised to explore the action space which has the maximum disagreement or variance between the predictions of an ensemble of forward-dynamics models. The models converges to mean, which reduces the variance of the ensemble and prevents it from getting stuck in stochasticity traps.

\textbf{Coverage}. In the \emph{coverage} approach, the agent tries to maximise the amount of targets it directly observes. Typically, this would be the area seen in an environment \cite{chen2019learning, chaplot2020learning, ramakrishnan2020exploration}. Since the agent uses egocentric observations, it has to navigate based on possibly obstructive 3D structures. One recent method combines classic and learning-based methods \cite{chaplot2020learning}. It uses analytical path planners with a learned SLAM module that maintains a spatial map, to avoid the high sample complexities involved in training end-to-end policies. This method also includes noise models to improve physical realism for generalisability to real-world robotics. Another recent work is a scene memory transformer which uses the self-attention mechanism adapted from the Transformer model \cite{vaswani2017attention} over the scene memory in its policy network \cite{fang2019scene}. The scene memory embeds and stores all encountered observations, allowing for greater flexibility and scalability as compared to a map-like memory that requires inductive biases.

\textbf{Reconstruction}. In the \emph{reconstruction} approach, the agent tries to recreate other views from an observed view. Past work focuses on pixel-wise reconstructions of 360 degree panoramas and CAD models \cite{ramakrishnan2018sidekick, song2018im2pano3d}, which are usually curated datasets of human-taken photos \cite{ramakrishnan2020occupancy}. Recent work has adapted this approach for embodied AI, which is more complex because the model has to perform scene reconstruction from the agent's egocentric observations and the control of its own sensors (i.e. active perception). In a recent work, the agent uses its egocentric RGB-D observations to reconstruct the occupancy state beyond visible regions and aggregate its predictions over time to form an accurate occupancy map \cite{ramakrishnan2020occupancy}. The occupancy anticipation is a pixel-wise classification task where each cell in a local area of V x V cells in front of the camera is assigned probabilities of it being explored and occupied. As compared to the \emph{coverage} approach, anticipating the occupancy state allows the agent to deal with regions that are not directly observable. Another recent work focuses on semantic reconstruction rather than pixel-wise reconstruction \cite{ramakrishnan2020exploration}. The agent is designed to predict whether semantic concepts like ``door" are present at sampled query locations. Using a \emph{K}-means approach, the true reconstruction concepts for a query location are the \emph{J} nearest cluster centroids to its feature representation. The agent is rewarded if it obtains views that help it predict the true reconstruction concepts for sampled query views.

\subsubsection{Evaluation Metrics}
Amount of targets visited. Different types of targets are considered, such as area \cite{chaplot2020learning, savinov2018episodic} and interesting objects \cite{fang2019scene, haber2018learning}. The area visited metric has a few variants, such as the absolute coverage area in $m^2$ and the percentage of the area explored in the scene.

\textbf{Impact on downstream tasks}. Visual exploration performance can also be measured by its impact on downstream tasks like visual navigation. This evaluation metric category is more commonly seen in recent works. Examples of downstream tasks that make use of visual exploration outputs (i.e. maps) include Image Navigation \cite{beeching2020learning, mezghani2020learning}, Point Navigation \cite{chaplot2020learning, anderson2018evaluation} and Object Navigation \cite{wahid2020learning, du2020learning, chaplot2020object}. More details about these navigation tasks can be found in section \ref{navigation}.

\subsubsection{Datasets}
For visual exploration, some popular datasets include Matterport3D and Gibson V1. Matterport3D and Gibson V1 are both photorealistic RGB datasets with useful information for embodied AI like depth and semantic segmentations. The Habitat-Sim simulator allows for the usage of these datasets with extra functionalities like configurable agents and multiple sensors. Gibson V1 has also been enhanced with features like interactions and realistic robot control to form iGibson. However, more recent 3D simulators like those mentioned in section \ref{sec:simulator} can all be used for visual exploration, since they all offer RGB observations at the very least.

\subsection{Visual Navigation}\label{navigation}

In visual navigation, an agent navigates a 3D environment to a goal with or without external priors or natural language instruction. Many types of goals have been used for this task, such as points, objects, images \cite{zhu2017target, chaplot2020neural} and areas \cite{anderson2018evaluation}. We will focus on points and objects as goals for visual navigation in this paper, as they are the most common and fundamental goals. They can be further combined with specifications like perceptual inputs and language to build towards more complex visual navigation tasks, such as \emph{Navigation with Priors}, \emph{Vision-and-Language Navigation} and even \emph{Embodied QA}. Under point navigation \cite{ye2020auxiliary}, the agent is tasked to navigate to a specific point while in object navigation \cite{campari2020exploiting, batra2020objectnav}, the agent is tasked to navigate to an object of a specific class.

% Some observations that can be used apart from the basic RGB include audio \cite{chen2020audio} and depth... ***FIND CITATIONS***

% \textcolor{blue}{kiv}
While classic navigation approaches \cite{bonin2008visual} are usually composed of hand-engineered sub-components like localization, mapping \cite{fuentes2015visual}, path planning \cite{kavraki1996probabilistic, lavalle2001rapidly} and locomotion. Visual navigation in embodied AI aims to learn these navigation systems from data, so as to reduce case-specific hand-engineering, hence easing integration with downstream tasks having superior performance with the data-driven learning methods, such as question answering \cite{ye2020seeing}. There are also hybrid approaches \cite{chaplot2020learning} that aim to combine the best of both worlds. As previously mentioned in section \ref{sec:simulator}, learning-based approaches are more robust to sensor measurement noise as they use RGB and/or depth sensors and are able to incorporate semantic understanding of an environment. Furthermore, they enable an agent to generalize its knowledge of previously seen environments to help understand novel environments in an unsupervised manner, reducing human effort.

Along with the increase in research in recent years, challenges have also been organised for visual navigation in the fundamental point navigation and object navigation tasks to benchmark and accelerate progress in embodied AI \cite{batra2020objectnav}. The most notable challenges are the iGibson Sim2Real Challenge, Habitat Challenge \cite{habitat2020sim2real} and RoboTHOR Challenge. For each challenge, we will describe the 2020 version of the challenges, which is the latest as of this paper. In all three challenges, the agent is limited to egocentric RGB-D observations. For the iGibson Sim2Real Challenge 2020, the specific task is point navigation. 73 high-quality Gibson 3D scenes are used for training, while the Castro scene, the reconstruction of a real world apartment, will be used for training, development and testing. There are three scenarios: when the environment is free of obstacles, contains obstacles that the agent can interact with, and/or is populated with other moving agents. For the Habitat Challenge 2020, there are both point navigation and object navigation tasks. Gibson 3D scenes with Gibson dataset splits are used for the point navigation task, while 90 Matterport3D scenes with the 61/11/18 training/validation/test house splits specified by the original
dataset \cite{anderson2018evaluation, chang2017matterport3d} are used for the object navigation task. For the RoboTHOR Challenge 2020, there is only the object navigation task. The training and evaluation are split into three phases. In the first phase, the agent is trained on 60 simulated apartments and its performance is validated on 15 other simulated apartments. In the second phase, the agent will be evaluated on four simulated apartments and their real-world counterparts, to test its generalisation to the real world. In the last phase, the agent will be evaluated on 10 real-world apartments.

In this section, we build upon existing visual navigation survey papers \cite{anderson2018evaluation, mishkin2019benchmarking, ye2020seeing} to include more recent works.

\subsubsection{Categories}
\textbf{Point Navigation} has been one of the foundational and more popular tasks \cite{chaplot2020learning} in recent visual navigation literature. In point navigation, an agent is tasked to navigate to any position within a certain fixed distance from a specific point \cite{anderson2018evaluation}. Generally, the agent is initialized at the origin $(0,0,0)$ in an environment, and the fixed goal point is specified by 3D coordinates $(x,y,z)$ relative to the origin/initial location \cite{anderson2018evaluation}. For the task to be completed successfully, the artificial agent would need to possess a diverse range of skillsets such as visual perception, episodic memory construction, reasoning/planning, and navigation. The agent is usually equipped with a GPS and compass that allows it to access to their location coordinates, and implicitly their orientation relative to the goal position \cite{savva2019habitat, ye2020auxiliary}. The target's relative goal coordinates can either be static (i.e. given only once, at the beginning of the episode) or dynamic (i.e. given at every time-step) \cite{savva2019habitat}. More recently, with imperfect localization in indoor environments, Habitat Challenge 2020 has moved on to the more challenging task \cite{wijmans2019dd} of RGBD-based online localization without the GPS and compass.

There have been many learning-based approaches to point navigation in recent literature. One of the earlier works \cite{mishkin2019benchmarking} uses an end-to-end approach to tackle point navigation in a realistic autonomous navigation setting (i.e. unseen environment with no ground-truth maps and no ground-truth agent's poses) with different sensory inputs. The base navigation algorithm is the Direct Future Prediction (DFP) \cite{dosovitskiy2016learning} where relevant inputs such as color image, depth map and actions from the four most recent observations are processed by appropriate neural networks (e.g. convolutional networks for sensory inputs) and concatenated to be passed into a two-stream fully connected action-expectation network. The outputs are the future measurement predictions for all actions and future time steps. 

The authors also introduce the Belief DFP (BDFP), which is intended to make the DFP's black-box policy more interpretable by introducing an intermediate map-like representation in future measurement prediction. This is inspired by the attention mechanism in neural networks, and successor representations \cite{dayan1993improving, zhu2017visual} and features \cite{barreto2017successor} in reinforcement learning. Experiments show that the BDFP outperforms the DFP in most cases, classic navigation approaches generally outperform learning-based ones with RGB-D inputs. \cite{gordon2019splitnet} provides a more modular approach. For point navigation, SplitNet's architecture consists of one visual encoder and multiple decoders for different auxiliary tasks (e.g. egomotion prediction) and the policy. These decoders aim to learn meaningful representations. With the same PPO algorithm \cite{schulman1707proximal} and behavioral cloning training, SplitNet can outperform comparable end-to-end methods in previously unseen environments.

Another work presents a modular architecture for simultaneous mapping and target-driven navigation in indoors environments \cite{georgakis2019simultaneous}. In this work, the authors build upon MapNet \cite{henriques2018mapnet} to include 2.5D memory with semantically-informed features and train a LSTM for the navigation policy. They show that this method outperforms a learned LSTM policy without a map \cite{mousavian2019visual} in previously unseen environments.

With the introduction of the \emph{Habitat Challenge} in 2019 and its standardized evaluation, dataset and sensor setups, the more recent approaches have been evaluated with the \emph{Habitat Challenge 2019}. The first work comes from the team behind Habitat, and uses the PPO algorithm, the actor-critic model structure and a CNN for producing embeddings for visual inputs. A follow-up work provides an ``existence proof" that near-perfect results can be achieved for the point navigation task for agents with a GPS, a compass and huge learning steps (2.5 billion steps as compared to Habitat's first PPO work with 75 million steps) in unseen environments in simulations \cite{wijmans2019dd}. Specifically, the best agent's performance is within 3-5$\%$ of the shortest path oracle. This work uses a modified PPO with Generalized Advantage Estimation \cite{schulman2015high} algorithm that is suited for distributed reinforcement learning in resource-intensive simulated environments, namely the Decentralized Distributed Proximal Policy Optimization (DD-PPO). At every time-step, the agent receives an egocentric observation (depth or RGB), gets embeddings with a CNN, utilizes its GPS and compass to update the target position to be relative to its current position, then finally outputs the next action and an estimate of the value function. The experiments show that the agents continue to improve for a long time, and the results nearly match that of a shortest-path oracle. 

The next work aims to improve on this resource-intensive work by increasing sample and time efficiency with auxiliary tasks \cite{ye2020auxiliary}. Using the same DD-PPO baseline architecture from the previous work, this work adds three auxiliary tasks: action-conditional contrastive predictive coding (CPC|A) \cite{guo2018neural}, inverse dynamics \cite{pathak2017curiosity} and temporal distance estimation. The authors experiment with different ways of combining the representations. At 40 million frames, the best performing agent achieves the same performance as the previous work $5.5X$ faster and even has improved performance. The winner of the Habitat Challenge 2019 for both the RGB and the RGB-D tracks \cite{chaplot2020learning} provides a hybrid solution that combines both classic and learning-based approaches as end-to-end learning-based approaches are computationally expensive. This work incorporates learning in a modular fashion into a ``classic navigation pipeline", thus implicitly incorporating the knowledge of obstacle avoidance and control in low-level navigation. The architecture consists of a learned Neural SLAM module, a global policy, a local policy and an analytical path planner. The Neural SLAM module predicts a map and agent pose estimate using observations and sensors. The global policy always outputs the target coordinates as the long-term goal, which is converted to a short-term goal using the analytic path planner. Finally, a local policy is trained to navigate to this short-term goal. The modular design and use of analytical planning help to reduce the search space during training significantly.

\textbf{Object Navigation} is one of the most straightforward tasks, yet one of the most challenging tasks in embodied AI. Object navigation focuses on the fundamental idea of navigating to an object specified by its label in an unexplored environment \cite{batra2020objectnav}. The agent will be initialized at a random position and will be tasked to find an instance of an object category within that environment. Object navigation is generally more complex than point navigation, since it not only requires many of the same skillsets such as visual perception and episodic memory construction, but also semantic understanding. These are what makes the object navigation task much more challenging, but also rewarding to solve. 

%Original

The task of object navigation can be demonstrated or learnt through adapting, which helps to generalize navigation in an environment without any direct supervision. This work \cite{wortsman2019learning} achieves that through a meta-reinforcement learning approach, as the agent learns a self-supervised interaction loss which helps to encourage effective navigation. Unlike the conventional navigation approaches for which the agents freeze the learning model during inference, this work allows the agent learns to adapt itself in a self-supervised manner and adjust or correct its mistake afterwards. This approach prevents an agent from making too many mistakes before realizing and make the necessary correction. Another method is to learn the object relationship between objects before executing the planning of navigation. This work \cite{du2020learning} implements an object relation graph (ORG) which is not from external prior knowledge but rather a knowledge graph that is built during the visual exploration phase. The graph consists of object relationships such as category closeness and spatial correlations.

\textbf{Navigation with Priors} focuses on the idea of injecting semantic knowledge or priors in the form of multimodal inputs such as knowledge graph or audio input or to aid in the training of navigation tasks for embodied AI agents in both seen and unseen environments. Past work \cite{yang2018visual} that use human priors of knowledge integrated into a deep reinforcement learning framework has shown that artificial agent can tap onto human-like semantic/functional priors to aid the agent in learning to navigate and find unseen objects in the unseen environment. Such example taps onto the understanding that the items of interest, such as finding an apple in the kitchen, humans will tend to look at logical locations to begin our search. These knowledge are encoded in a graph network and trained upon in a deep reinforcement learning framework.

There are other examples of using human priors such as human's ability to perceive and capture correspondences between an audio signal modal and the physical location of objects hence to perform navigation to the source of the signal. In this work \cite{gan2020look}, artificial agents pick multiple sensory observations such as vision and sound signal of the target objects and figure out the shortest trajectory to navigation from its starting location to the source of the sounds. This work achieves it through having a visual perception mapper, sound perception module and dynamic path planners.

\textbf{Vision-and-Language Navigation} (VLN) is a task where agents learn to navigate the environment by following natural language instructions. The challenging aspect of this task is to perceive both the visual scene and language sequentially. VLN remains a challenging task as it requires agents to make predictions of future actions based on past actions and instructions \cite{anderson2018evaluation}. Furthermore, agents might not be able to align their trajectories seamlessly with natural language instructions. Although vision-and-language navigation and visual question answering (VQA) might seem similar, there are major differences in both tasks. Both tasks can be formulated as visually grounded, sequence-to-sequence transcoding problems. However, VLN sequences are much longer and require vision data to be constantly fed as input and the ability to manipulate camera viewpoints, as compared to VQA where a single input question is fed in and an answer is generated. We are now able to give a natural language instruction to a robot and expect them to perform the task \cite{anderson2018vision,duan2020actionet,shridhar2020alfred}. These are achieved with the advancement of recurrent neural network methods \cite{anderson2018vision} for joint interpretation of both visual and natural language inputs and datasets that are designed for simplifying processes of task-based instruction in navigation and performing of tasks in the 3D environment.  

One approach for VLN is the Auxiliary Reasoning Navigation framework \cite{zhu2020vision}.
It tackles four auxiliary reasoning tasks: trajectory retelling, progress estimation, angle prediction and cross-modal matching. The agent learns to reason about the previous actions and predicts future information the tasks.

Vision-dialog navigation is the latest extension of VLN as it aims to train an agent to develop the ability to engage in a constant natural language conversation with humans to aid in its navigation. The current work \cite{zhu2020vision2} in this area uses a Cross-modal Memory Network (CMN) that remembers and understands useful information related to past navigation actions through separate language memory and visual memory modules, and further uses it to make decisions for navigation.

\subsubsection{Evaluation Metrics}
Visual navigation uses (1) success
weighted by (normalized inverse) path length (SPL) and (2) success rate as the main evaluation metrics \cite{anderson2018evaluation}. Success weighted by path length can be defined as: {\large $\frac{1}{N}\sum^{N}_{i=1} S_i \frac{l_i}{max(p_i,l_i)}$}. $S_i$ is a success indicator for episode $i$, $p_i$ is the agent's path length, $l_i$ is the shortest path length and $N$ is the number of episodes. It is noteworthy that there are some known issues with success weighted by path length \cite{batra2020objectnav}. Success rate is the fraction of the episodes in which the agent reaches the goal within the time budget \cite{mishkin2019benchmarking}. There are also other less common evaluation metrics \cite{anderson2018evaluation, mishkin2019benchmarking, georgakis2019simultaneous, campari2020exploiting, chaplot2020object} in addition to the two mentioned, namely: (3) path length ratio, which is the ratio between the predicted path and the shortest path length and is calculated only for successful episodes; (4) distance to success/navigation error, which measures the distance between the agent's final position and the success threshold boundary around the nearest object or the goal location respectively.

Besides the above four metrics, there are another two metrics used to evaluate VLN agents. They are: (1) oracle success rate, the rate for which the agent stops at the closest point to the goal along its trajectory; (2) trajectory length. In general, for VLN tasks, the best metric is still SPL as it takes into account of the path taken and not just the goal.

For vision-dialog navigation, in addition to success rate and oracle success rate, there are another two metrics used: (1) goal progress, the average agent progress towards the goal location; (2) oracle path success rate, the success rate of agent stopping at the closest point to goal along the shortest path.

\subsubsection{Datasets}

As in visual exploration, Matterport3D and Gibson V1 are the most popular datasets. It is noteworthy that the scenes in Gibson V1 are smaller and usually have shorter episodes (lower GDSP from start position to goal position). The AI2-THOR simulator/dataset is also used.

Unlike the rest of the visual navigation tasks, VLN requires a different kind of dataset. Most of the VLN works use the Room-to-Room (R2R) dataset with the Matterport3D Simulator \cite{mattersim}. It consists of 21,567 navigation instructions with an average length of 29 words. In vision-dialog navigation \cite{zhu2020vision}, the Cooperative Vision-and-Dialog Navigation (CVDN) \cite{thomason:corl19} dataset is used. It comprises 2,050 human-to-human dialogs and over 7,000 trajectories within the Matterport3D Simulator.

\subsection{Embodied Question Answering}

The task of embodied question answering (QA) in recent embodied AI simulators has been a significant advancement in the field of general-purpose intelligence systems. To perform QA in a state of physical embodiment, an AI agent would need to possess a wide range of AI capabilities such as visual recognition, language understanding, question answering, commonsense reasoning, task planning, and goal-driven navigation. Hence, embodied QA can be considered the most onerous and complicated task in embodied AI research currently.  

% \textcolor{blue}{why is this section dataset first?}

\subsubsection{Categories}
% \textbf{Embodied Question Answering (EmbodiedQA) is a QA task that requires an agent to interact with a dynamic visual environment.  This task is also relevant to vision-language grounding~\cite{Huang_2018_CVPR,mao2016generation}. 

For \textbf{embodied QA} (EQA), a common framework that divides the task into two sub-tasks: a navigation task and a QA task. The navigation module is essential since the agent needs to explore the environment to see the objects before answering questions about them. For example, \cite{das2018embodied} proposed the Planner-Controller Navigation Module (PACMAN), which comprises a hierarchical structure for the navigation module, with a planner that selects actions (directions) and a controller that decides how far to move following each action. Once the agent decide to stop, the QA module is executed by using the sequence of frames along its path. The navigation module and visual question answering module are first trained individually and then jointly trained by REINFORCE \cite{williams1992simple}. \cite{das2018neural} and \cite{MultiTarget;YuCGBBB19} further improved the PACMAN model with the Neural Modular Control (NMC) where the higher-level master policy proposes semantic sub-goals to be executed by sub-policies.

% All these modules are all pre-trained separately. The question and answering accuracy and the average episode length were taken to answer the questions were used as the evaluation metrics, and the percentage of invalid action is also evaluated.  Since this task of Embodied Question and Answering is relatively new, hence most of the benchmark results were compared internally. 

\textbf{Multi-target embodied QA} (MT-EQA) \cite{MultiTarget;YuCGBBB19} is a more complex embodied QA task, which studies questions that have multiple targets in them, e.g. "Is the apple in the bedroom bigger than the orange in the living room?", such that the agent has to navigate to the "bedroom" and the "living room" to localize the "apple" and the "orange" and then perform comparisons to answer the questions. 

\textbf{Interactive Question Answering} (IQA) \cite{gordon2018iqa} is another work tackling the task of embodied QA in the AI2-THOR environment. IQA is an extension of EQA because it is essential for the agent to interact with the objects to answer certain questions successfully (e.g. the agent needs to open the refrigerator to answer the existence question "Is there an egg in the fridge?"). \cite{gordon2018iqa} proposed using a Hierarchical Interactive Memory Network (HIMN), which is a hierarchy of controllers that help the system operate, learn and reason across multiple time scales, while simultaneously reducing the complexity of each sub-task. An Egocentric Spatial Gated Recurrent Unit (GRU) acts as a memory unit for retaining spatial and semantic information of the environment. The planner module will have control over the other modules such as a navigator which runs an A* search to find the shortest path to the goal, a scanner which performs rotation for detecting new images, a manipulator that is invoked to carry out actions to change the state of the environment and lastly an answerer that will answer the question posted to the AI agent. \cite{MultiAgentVQA;TanXLGS20} studied IQA from a multi-agent perspective, where several agents explore an interactive scene jointly to answer a question. \cite{MultiAgentVQA;TanXLGS20} proposed multi-layer structural and semantic memories as scene memories to be shared by multiple agents to first reconstruct the 3D scenes and then perform QA.

\subsubsection{Evaluation Metrics}
Embodied QA and IQA involve two sub-tasks: \textit{1) navigation}, and \textit{2) question answering}, and these two sub-tasks are evaluated based on different metrics.

Navigation performance is evaluated by: (1) distance to target at navigation termination, i.e. navigation error ($d_T$); (2) change in distance to target from initial to final position, i.e. goal progress ($d_\Delta$); (3) smallest distance to target at any point in the episode ($d_{min}$); (4) percentage of episodes agent terminates navigation for answering before reaching the maximum episode length ($\%stop$); (5) percentage of questions where the agent terminates in the room containing the target object ($\%r_T$); (6) percentage of questions where the agent enters the room containing the target object at least once ($\%r_e$); (7) Intersection over Union for target object (IoU); (8) hit accuracy based on IoU ($h_T$); (9) episode length, i.e. trajectory length. Metrics (1), (2) and (9) are also used as evaluation metrics for the visual navigation task.

QA performance is evaluated by: (1) mean rank (MR) of the ground-truth answer in predictions; (2) accuracy. 
%\end{itemize}

% Question Answering Accuracy. Our agent (and all baselines) produce a probability distribution over 172 possible
% answers (colors, rooms, objects). We report the mean rank
% (MR) of the ground-truth answer in the answer list sorted
% by the agent’s beliefs, where the mean is computed over all
% test questions and environments.

% Navigation Accuracy. We evaluate navigation performance on EQA v1 by reporting the distance to the target object at navigation termination pdTq, change in distance to target from initial to final position pd∆q, and the smallest
% distance to the target at any point in the episode pdminq. All
% distances are measured in meters along the shortest path to
% the target. We also record the percentage of questions for
% which an agent either terminates in p%rTq or ever enters
% p%rŒq the room containing the target object(s). Finally, we
% also report the percent of episodes in which agents choose to
% terminate navigation and answer before reaching the maximum episode length p%stopq. To sweep the difficulty of
% the task at test time, we spawn the agent 10, 30, or 50 actions away from the target and report each metric for T´10,
% T´30, T´50 settings.

\subsubsection{Datasets}
The EQA \cite{das2018embodied} dataset is based on House3D, a subset of the popular SUNCG \cite{song2016ssc} dataset with synthesized rooms and layouts that is similar to the Replica dataset \cite{straub2019replica}. House3D converts SUNCG's static environment into a virtual environment, where the agent can navigate with physical constraints (e.g. it cannot pass through walls or objects). To test the agent's capabilities in language grounding, commonsense reasoning and navigation, \cite{das2018embodied} uses a series of functional programs in CLEVR \cite{johnson2017clevr} to synthesize questions and answers regarding objects and their properties (e.g. color, existence, location and relative preposition). In total, there are 5,000 questions in 750 environments with reference to 45 unique objects in 7 unique room types. 

For MT-EQA \cite{MultiTarget;YuCGBBB19}, the authors introduce the MT-EQA dataset, which contains 6 types of compositional questions which compare object attribute properties (color, size, distance) between multiple targets (objects/rooms).

For IQA \cite{gordon2018iqa}, the authors annotated a large scale dataset, IQUAD V1, which consist of 75,000 multiple-choice questions. Similar to the EQA dataset, IQUAD V1 has questions regarding object existence, counting and spatial relationships.

\begin{figure*}[thb]
\begin{minipage}[b]{1.0\linewidth}
  \centering
  \centerline{\includegraphics[width=\linewidth]{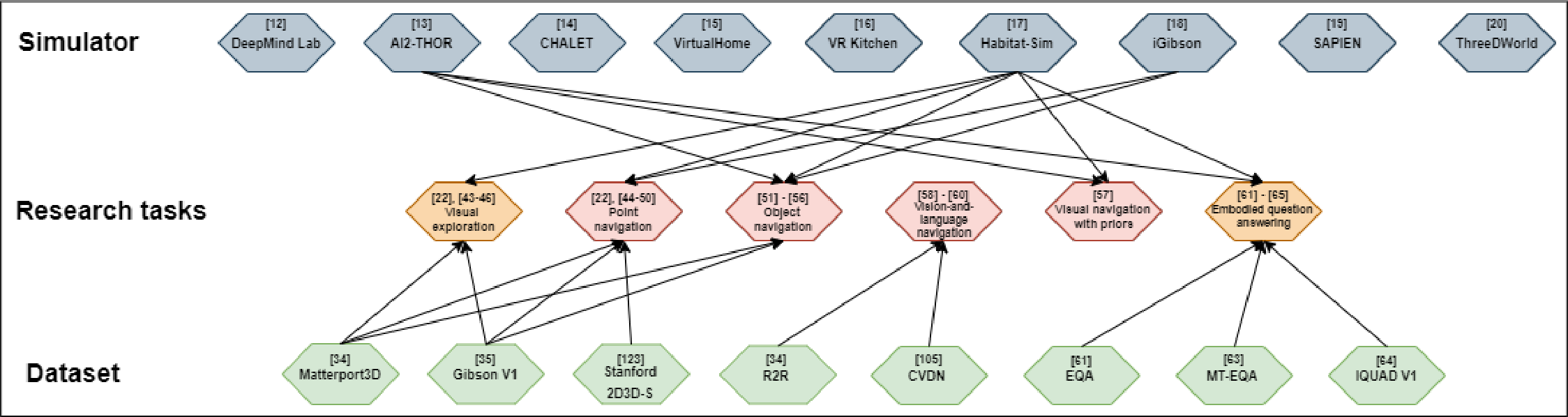}}
  \caption{Connections between Embodied AI simulators to research. (Top) Nine up-to-date embodied AI simulators. (Middle) The various embodied AI research tasks as a result of the nine embodied AI simulators. The red colored research tasks are grouped under the visual navigation category while the rest of the yellow colored tasks are the other research categories. (Bottom) The evaluation dataset used in the evaluation of the research tasks in one of the nine embodied AI simulators..}
  
%   \textcolor{blue}{still have the following and previous issues. (1) use same form. use "dataset" or "Dataset" consistently - EQA "dataset" (2) If possible, include the year and reference of dataset. (3) vln is visual-and-language navigation below} \textcolor{red}{Changed}
  
\label{fig:connection}
\end{minipage}
\end{figure*}

\section{Insights and Challenges}
\label{challenges}

\subsection{Insights into Embodied AI}
The interconnections in Fig. \ref{fig:connection} reflects the suitability of simulators to research tasks. Based on Fig. \ref{fig:connection}, both Habitat-Sim and iGibson support research tasks in visual exploration and a range of visual navigation tasks, indicating the importance of high fidelity, which comes from \emph {world-based scene} simulators. However, because of their distinct unique features that make them preferable for non-embodied AI standalone tasks such as in deep reinforcement learning, some simulators do not presently connect to any of the embodied research tasks. Nonetheless, they still meet the criteria for being classified as embodied AI simulators.

%original

On the contrary, research tasks such as embodied question answering and visual navigation with priors would require the embodied AI simulators to have \emph{multiple-state} object property, due to the interactive nature of these tasks. Hence, AI2-THOR is undoubtedly the simulator of choice. Lastly, VLN is the only research task that currently does not utilize any of the nine embodied AI simulators but instead uses Matterport3D Simulator \cite{mattersim}. This is because previous works in VLN does not require the feature of \emph{interactivity} in its simulator; hence Matterport3D simulator suffice. However, with the furtherance of VLN tasks, we can expect the need for interactions in VLN tasks, hence the need to use embodied AI simulators. Furthermore, unlike traditional reinforcement learning simulation environments \cite{todorov2012mujoco,brockman2016openai} focus on task specific training, while embodied AI simulators provide a training environment for training a wide range of different tasks akin to those undertaken in the physical world.

% \textcolor{red}{Last point is still unclear. Why is last task not using the 9? Any feature missing?}

Furthermore, based on the survey done on the embodied AI research tasks in section \ref{sec:research}, we propose a pyramid structure in which each embodied AI research task contributes to the next. Visual exploration, for example, aids in the development of visual navigation, and visual navigation contributes to the creation of embodied QA. This build-up approach also correlates with the increasing complexity of the tasks.
Based on the foreseeable trends in embodied AI research, we hypothesize that a next advancement in the pyramid of embodied AI research is \textbf{Task-based Interactive Question Answering (TIQA)}, which aims to integrate tasks with answering specific questions. For example, such questions can be ``\emph{How long would it take for an egg to boil? Is there an apple in the cabinet?}". These are questions that cannot be answered through the conventional approaches \cite{das2018embodied,gordon2018iqa}.
They require the embodied agent to perform specific tasks related to the questions to unlock new insights that are momentous in answering those QA questions. The \textbf{TIQA} agents that we hypothesize can perform an array of general household tasks, which allows them to extrapolate useful environmental information that is crucial in helping them to derive the answer to the QA questions. TIQA may hold the key to generalizing task-planning and developing general-purpose AI in simulations which later can be deployed into the real world.

\subsection{Challenges in Embodied AI Simulators}
Current embodied AI simulators have reached a level in both its functionality and fidelity, that sets them apart from those conventional simulation used for reinforcement learning. Even with this soaring variance of embodied AI simulators, there are several existing challenges in embodied AI simulators in areas ranging from their \textbf{realism}, \textbf{scalability} to \textbf{interactivity}.
%These challenges are closely related to our proposed second level comparison metrics. 

\textbf{Realism}. It focuses on the \emph{fidelity} and \emph{physics} features of the simulators. Simulators with both a high visual fidelity and realistic physics are highly sought after by the robotics communities as they provide the ideal test-bed for various robotic tasks such as navigation and interaction tasks \cite{gao2021room,sun2021plate}. However, there is a lack of embodied AI simulators that possess both of \emph{world-based scene} and \emph{advanced physics}.

For \emph{fidelity}, simulators that are \emph{world-based scene} will undoubtedly outperform\emph{game-based scene} simulator in simulation to real tasks \cite{sadeghi2016cad2rl,kadian2020sim2real}. Despite this observation, only Habitat-Sim \cite{savva2019habitat}, and iGibson \cite{xia2020interactive} are \emph{world-based scene} simulators. This paucity of \emph{world-based scene} simulators is the bottleneck to simulation-to-real tasks for embodied AI agents, which further hinders the transferability of embodied AI research into real-world deployment. For \emph{physics}, the furtherance of physics-based predictive models \cite{bear2021physion,duan2021space,duan2021pip} have accentuate on the importance of embodied AI simulators with \emph{advanced physics features} as they serve to provide an ideal testbed for training embodied AI agents to perform tasks with sophisticated physical interactions \cite{duan2020actionet,shridhar2020alfred,nagarajan2020learning}. Despite the need for an advanced physics-based embodied AI simulator, there is currently only one simulator, ThreeDWorld \cite{gan2020threedworld} that fits this criterion. Hence, there is a severe lack of embodied AI simulators with advanced physics features such as cloth, fluid and soft-body physics. We believe that advances in 3D reconstruction techniques and physics engines \cite{wang2021thin,kuznetsov2021neumip,richter2021enhancing} will improve the realism of embodied AI. 

%Orgainal
% We believe that the realism of embodied AI have a quantum leap forward driven by the continuous advancement in 3D reconstruction techniques and advanced physics engines \cite{wang2021thin,kuznetsov2021neumip,richter2021enhancing}.

% \cite{groth2018shapestacks,Baradel_2020_ICLR,bear2021physion,duan2021space,duan2021pip}

% \cite{lohmann2020learning,duan2020actionet,shridhar2020alfred,nagarajan2020learning,khz2021interact}

% \textcolor{red}{This line seems incomplete. whats next? Improvement of world-base scene simulators to have more advanced physics features? or Improvement of advanced physics engines to have higher realism is still required to drive the next phases in embodied AI research tasks?} 

% \cite{chang2017matterport3d,xiazamirhe2018gibsonenv,uy-scanobjectnn-iccv19,replica19arxiv,ramakrishnan2021habitatmatterport}.

\textbf{Scalability}. Unlike image-based datasets \cite{deng2009imagenet,lin2014microsoft} which can be easily obtained from crowd-sourcing or the internet. The methodologies and tools are scarce for collecting large-scale world-based 3D scene datasets and 3D object assets \cite{replica19arxiv,ramakrishnan2021habitatmatterport,armeni2017joint}. These 3D scene datasets are crucial for the construction of a diverse of embodied AI simulators. Current approaches to collect realistic 3D scene datasets requires scanning of the physical room through photogrammetry \cite{mikhail2001introduction} such as Matterport 3D scanner, Meshroom \cite{Meshroom}, or even mobile 3D scanning applications. However, they are not commercially viable for collecting large scale 3D objects and scene scans. This is largely due to 3D scanners that are used for photogrammetry are costly and non-accessible. As such, the bottleneck to scalability lies in developing tools for large scale collection of high fidelity 3D object or scene scans. Hopefully, with the further advancement of 3D learning-based approaches \cite{yu2021pixelnerf,martin2021nerf} that aims to render 3D object meshes from a single or few images or even through scene generation approach \cite{hao2021gancraft}, we will be able to scale up the collection process of large scale 3D datasets.

\textbf{Interactivity}. The ability to have fine-grained manipulative interactions with functional objects in the embodied AI simulators are crucial in replicating human-level interactions with real-world objects \cite{zhu2020dark}. Most \emph{game-based scene} simulators \cite{kolve2017ai2,gao2019vrkitchen,xiang2020sapien,gan2020threedworld} provides both fine-grained object manipulation capabilities and symbolic interaction capabilities (e.g. $<$Pulldown Object X on Y$>$ action) or simply a ‘point-and-select’. However, due to the nature of \emph{game-based scene} simulators, many research tasks performed in this environment will opt for its symbolic interaction capabilities as compared to fine-grained object manipulation \cite{shridhar2020alfred}, except for a few that utilize both \cite{duan2020actionet,lohmann2020learning}. 

%Orignial 

On the other end, the agents from \emph{world-based scene} simulators \cite{savva2019habitat,xia2020interactive} possess the ability for gross motor control instead of the symbolic  interaction capabilities. However, the object property of the objects within these simulators being largely \emph{interact-able} on the surface which allows for gross motor control but lacks the \emph{multi-state} object classes which is number of state changes that the object have. Hence, there is a need to strike a balance in both the object functionality in its object property and also the complexity of \emph{action} that the embodied AI agent can perform in the environment.

Undoubtedly, mainstream simulators such as AI2-THOR \cite{kolve2017ai2}, iGibson \cite{xia2020interactive}, and Habitat-Sim \cite{savva2019habitat} do provide an excellent environment for advancing the respective embodied AI research. However, they do have their strengths and limitations to be overcome. With developments in computer graphics and computer vision, and the introduction of innovative real-world datasets, real-to-sim domain adaptation is one of the clear routes for improving embodied AI simulators. The concept of real-to-sim revolves around capturing real-world information such as tactile perception \cite{bhirangi2021reskin}, human-level motor control \cite{smith2020constraining} and audio inputs \cite{chen2019audio} in addition to visual sensory inputs and integrating them for the development of more realistic embodied AI simulators that can effectively bridge the physical and virtual worlds.

\subsection{Challenges in Embodied AI Research}
\label{sec:researchchallenges}

Embodied AI research tasks mark an increase in complexity from ``internet AI" to autonomous embodied learning agents in 3D simulated environments with multiple sensor modalities and potentially long trajectories \cite{chang2017matterport3d, ramakrishnan2020exploration}. This has led to \textbf{memory} and internal representations of the agent becoming extremely important \cite{ramakrishnan2020exploration, anderson2018evaluation, wahid2020learning}. Long trajectories and multiple input types also signified the importance of robust memory architecture which allows the agent to focus on the important parts of its environment. In recent years, there has been many different types of memory used, such as recurrent neural networks \cite{wijmans2019dd, ye2020auxiliary, wortsman2019learning, wahid2020learning, anderson2018vision, das2018embodied, das2018neural, MultiTarget;YuCGBBB19}, attention-based memory architectures, \cite{fang2019scene, campari2020exploiting, zhu2020vision2}, anticipated occupancy maps \cite{ramakrishnan2020occupancy}, occupancy maps \cite{ramakrishnan2020exploration} and semantic maps \cite{chaplot2020semantic, georgakis2019simultaneous, narasimhan2020seeing, gordon2018iqa, MultiAgentVQA;TanXLGS20}, with some papers having overwhelming emphasis on the novelty of their memory architectures \cite{ramakrishnan2020occupancy, ramakrishnan2020exploration, fang2019scene, zhu2020vision2}. However, while recurrent neural networks are known to be limited in capturing long-term dependencies in embodied AI \cite{fang2019scene, wahid2020learning}, it is currently still hard to agree which memory type(s) are better \cite{anderson2018evaluation} due to the lack of work focusing on memory architectures.

Among embodied AI research tasks, there has also been an increase in complexity, as seen in the progression from visual exploration to VLN and embodied QA where new components like language understanding and QA are added respectively. Each new component leads to exponentially harder and longer training of AI agents, especially since current approaches are often fully learning-based. This phenomenon has led to two promising advancements to reduce the search space and sample complexity while improving robustness: \textbf{hybrid approaches} combining classic and learning-based algorithms \cite{mishkin2019benchmarking, chaplot2020learning} and \textbf{prior knowledge incorporation} \cite{ye2020seeing, yang2018visual}. Furthermore, \textbf{ablation studies are much harder to manage} \cite{AllenAct} for more complex tasks as each new component in embodied AI makes it much harder to test for its contribution to the agent's performance, since it is added onto an existing set of components, and embodied AI simulators vary significantly in features and issues. This is compounded by the fact that research tasks have also increased in number rapidly. As a result, while some fundamental tasks like visual exploration have received more attention and thus have more approaches tackling them, the newer and more niche tasks like MT-EQA are much less addressed. New tasks usually introduce new considerations in important aspects like methods, evaluation metrics \cite{ramakrishnan2020exploration}, input types and model components, shown in Table \ref{tasks}, thus requiring even more evaluation than simpler tasks like visual exploration.

Lastly, there is a lack of focus on \textbf{multi-agent set-ups}, which contribute useful new tasks \cite{MultiAgentVQA;TanXLGS20}. This lack of focus can be attributed to the lack of simulators with multi-agent features until recently. Multi-agent systems for collaboration and communication are prevalent in the real world \cite{panait2005cooperative, liu2020who2com} but currently receive relatively little attention \cite{AllenAct}. With an increase in simulators with multi-agent features \cite{kolve2017ai2, shenigibson, gan2020threedworld} recently, it remains to be seen whether the multi-agent support (e.g. support for multi-agent algorithms) is sufficient.

\section*{Conclusion}
Recent advances in embodied AI simulators have been a key driver of progress in embodied AI research. 
Aiming to understand the trends and gaps in embodied AI simulators and research, this paper provides a contemporary and comprehensive overview of embodied AI simulators and research. The paper surveys nine embodied AI simulators and their connections in serving and driving recent innovations in research tasks for embodied AI. By benchmarking nine embodied AI simulators in terms of seven features, we seek to understand their provision of realism, scalability and interactivity, and hence use in embodied AI research. The three main tasks supporting the pyramid of embodied AI research -- visual exploration, visual navigation and embodied QA, are examined in terms of their approaches, evaluation metrics, and datasets. This is to review and benchmark the existing approaches in tackling these categories of embodied AI research tasks in the various embodied AI simulators. Furthermore, this paper allows us to unveil insightful relations between the simulators, datasets, and research tasks.
With the aid of this paper, AI researchers new to this field would be able to select the most suitable embodied AI simulators for their research tasks and contribute back to advancing the field of embodied AI.

\bibliographystyle{IEEEtran}
\bibliography{IEEEexample.bib}

 

\section{Biography Section}
% \vspace{-70pt}
\begin{IEEEbiography}[{\includegraphics[width=1in,height=1.25in,clip,keepaspectratio]{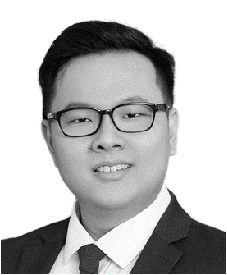}}]{Jiafei Duan}
received the B.Eng. (Highest Distinction) degree from the
School of Electrical and Electronics Engineering,
Nanyang Technological University,
Singapore, in 2021. He is currently working as a research engineer at the Institute of Infocomm Research, Agency for Science, Technology and Research (A*STAR), Singapore. His current research interest is
embodied AI and computational cognitive science.
\end{IEEEbiography}

\begin{IEEEbiography}[{\includegraphics[width=1in,height=1.25in,clip,keepaspectratio]{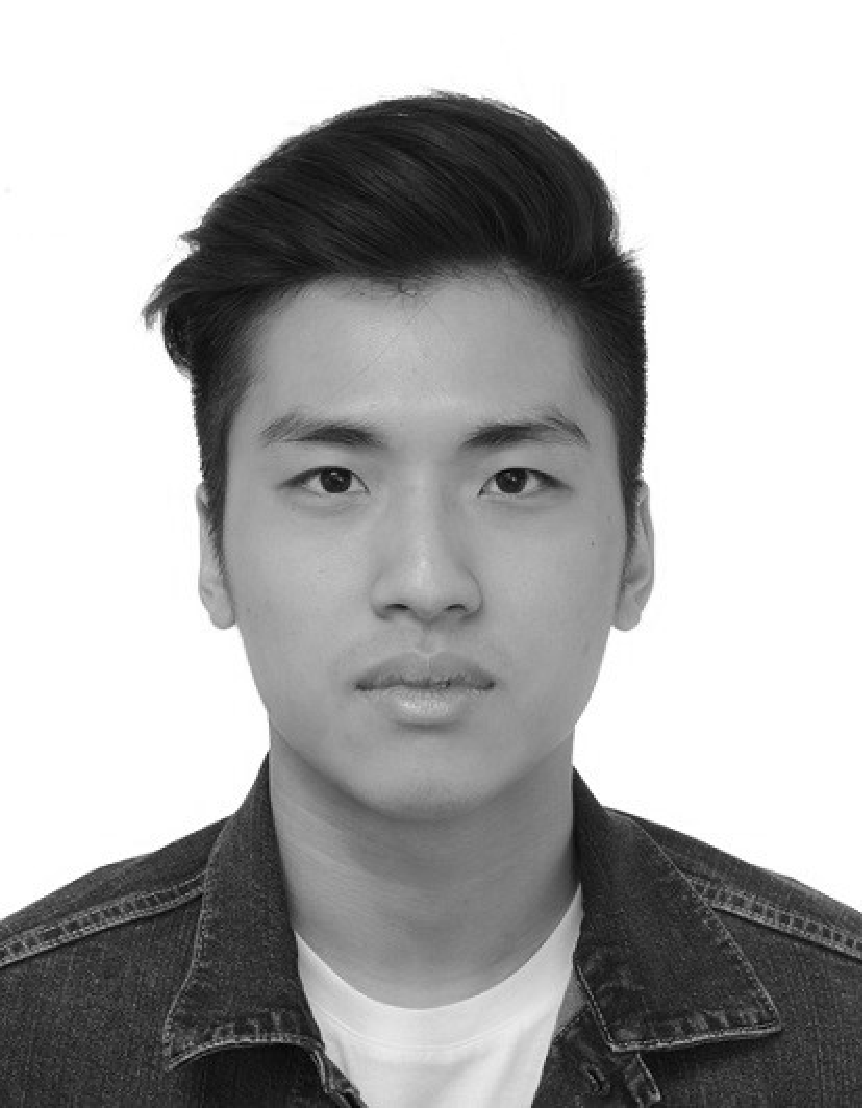}}]{Samson Yu}
received a B.Eng. degree in Information Systems Technology and Design from the Singapore University of Technology and Design in 2020. He is currently working as a research engineer at the Institute of High Performance Computing, Agency for Science, Technology and Research (A*STAR), Singapore, on fundamental AI research and embodied AI.
\end{IEEEbiography}

\begin{IEEEbiography}[{\includegraphics[width=1in,height=1.25in,clip,keepaspectratio]{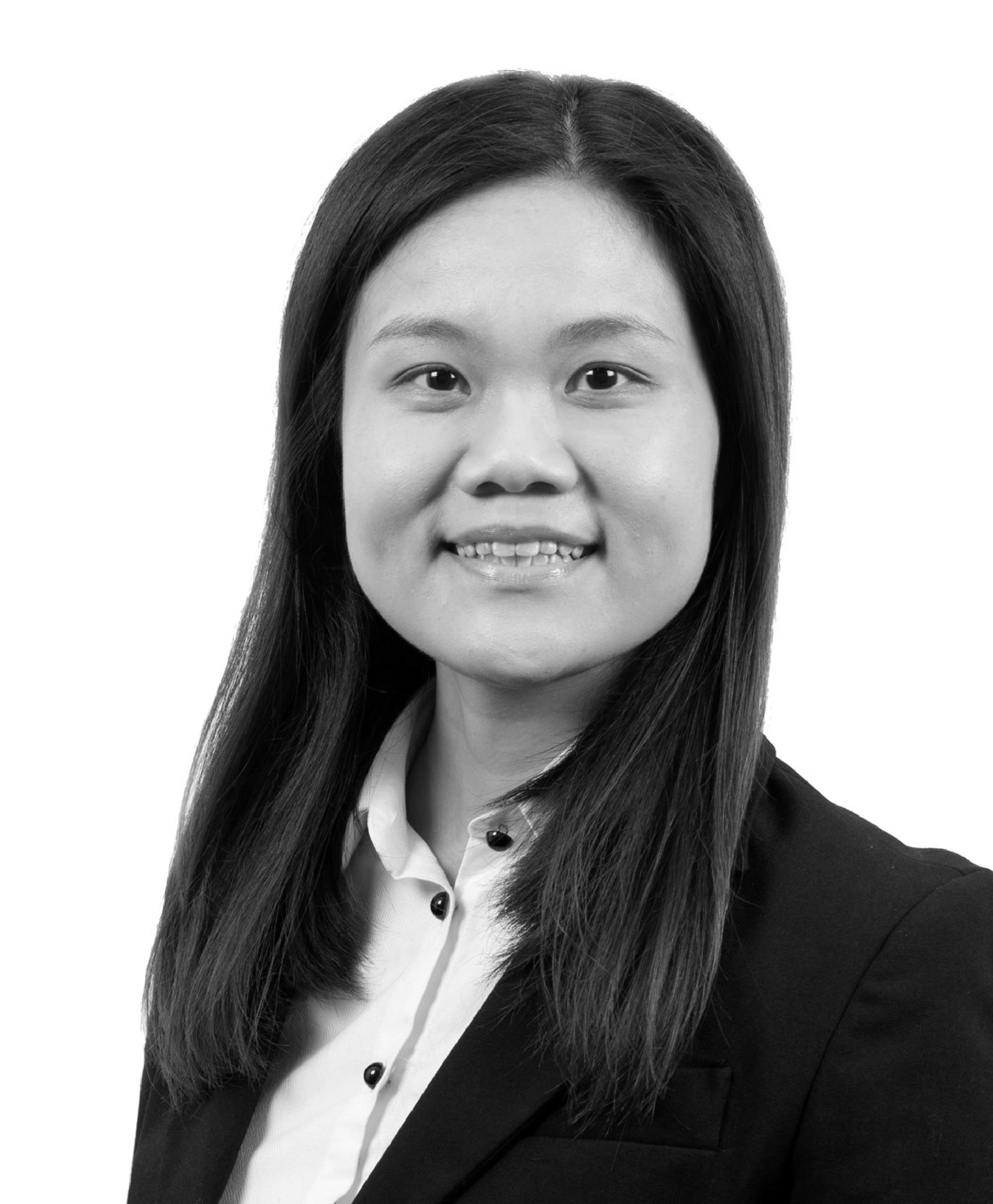}}]{Hui Li Tan} received the B.Sc. degree in applied mathematics from the National University of Singapore (NUS), Singapore, in 2007. She received  the Ph.D. degree in electrical and  computer engineering, from NUS in 2017. Since 2007, she has been with the Institute for Infocomm Research, Singapore. Her current research interests include computer vision, multimodal deep learning, incremental and federated learning.
\end{IEEEbiography}

\begin{IEEEbiography}[{\includegraphics[width=1in,height=1.25in,clip,keepaspectratio]{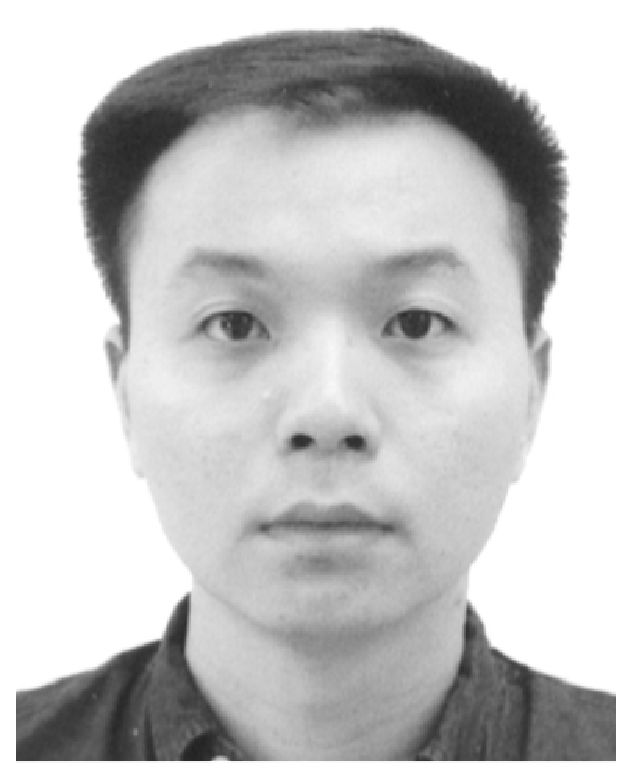}}]{Hongyuan Zhu}
received his Ph.D. degree in com-
puter  engineering  from  Nanyang  Technologi-
cal  University,  Singapore,  in  2014.  He  is  cur-
rently a Research Scientist with the Institute for
Infocomm  Research,  A*STAR,  Singapore.  His
research  interests  include  multimedia  content
analysis and segmentation.
\end{IEEEbiography}

\begin{IEEEbiography}[{\includegraphics[width=1in,height=1.25in,clip,keepaspectratio]{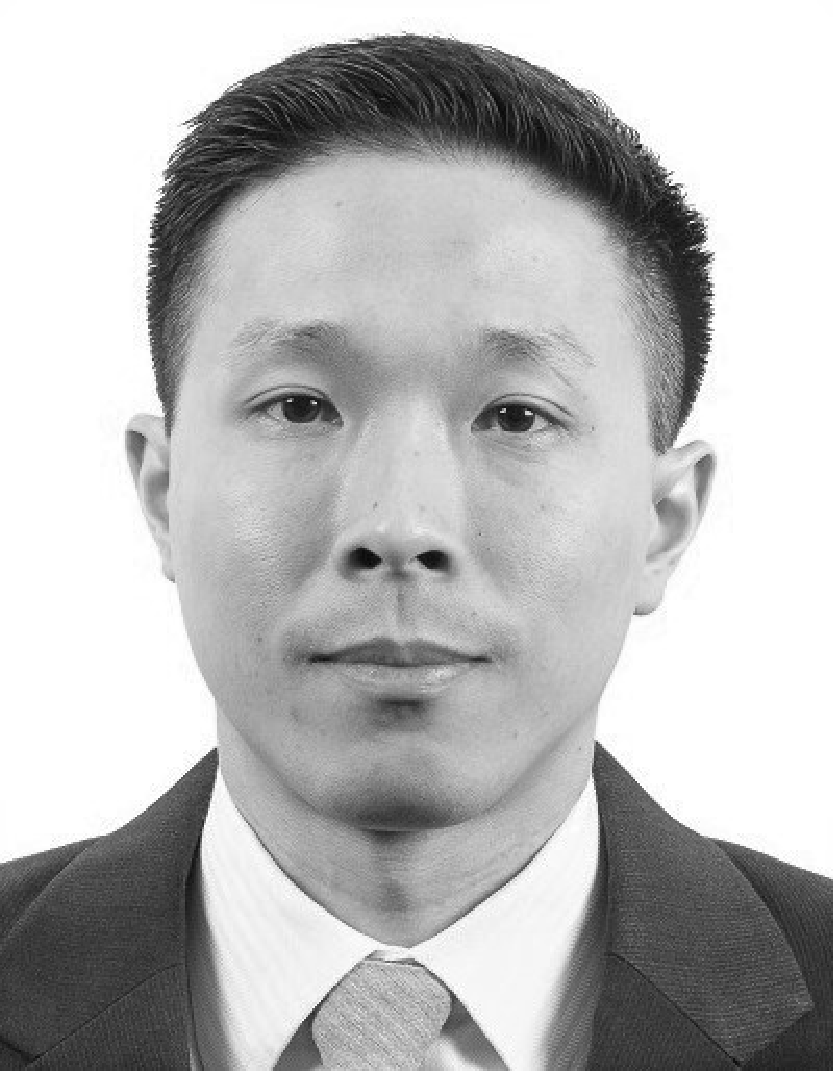}}]{Cheston Tan}
received B.Sc. (Highest Honours) degree from the Department of Electrical Engineering and Computer Science, University of California, Berkeley, as well as the Ph.D. degree from the Department of Brain and Cognitive Sciences, Massachusetts Institute of Technology. He is currently a senior scientist at the Institute for Infocomm Research (I2R), Agency for Science, Technology and Research (A*STAR), Singapore.
\end{IEEEbiography}



\vfill

\end{document}